\documentclass[11pt,a4paper]{article}
\usepackage[utf8]{inputenc}
\usepackage[T1]{fontenc}
\usepackage{lmodern}
\usepackage[margin=2.5cm]{geometry}
\usepackage{graphicx}
\usepackage{float}
\usepackage{amsmath,amssymb}
\usepackage{booktabs}
\usepackage{array}
\usepackage{authblk}
\usepackage{textcomp}
\usepackage{tipa}
\usepackage{caption}
\usepackage[colorlinks=true,allcolors=blue]{hyperref}

\title{Training-Free Cross-Lingual Dysarthria Severity Assessment via Phonological Subspace Analysis in Self-Supervised Speech Representations}

\author[1]{Bernard Muller}
\author[2]{Antonio Armando Ortiz Barra\~n\'on}
\author[1]{LaVonne Roberts\thanks{Corresponding author: lavonne@scottmorganfoundation.org}}
\affil[1]{The Scott-Morgan Foundation, Torquay, United Kingdom}
\affil[2]{Tecnol\'ogico de Monterrey, Monterrey, Mexico}
\date{}

\begin{document}
\maketitle

\noindent Funding: The Scott-Morgan Foundation \\
Conflicts of Interest: None declared

\section*{Abstract}
Dysarthric speech severity assessment typically requires either trained clinicians or supervised machine learning models built from labelled pathological speech data, limiting scalability across languages and clinical settings. We present a training-free method (no supervised severity model is trained; feature directions are estimated from healthy control speech using a pretrained forced aligner) that quantifies dysarthria severity by measuring the degradation of phonological feature subspaces within frozen HuBERT representations. For each speaker, we extract phone-level embeddings via Montreal Forced Aligner, compute $d'$ scores along phonological contrast directions (nasality, voicing, stridency, sonorance, manner, and four vowel features) derived exclusively from healthy control speech, and construct a 12-dimensional phonological profile. Evaluating 890 speakers across 10 corpora, 5 languages for the full MFA pipeline (English, Spanish, Dutch, Mandarin, French) and 3 primary aetiologies (Parkinson's disease, cerebral palsy, amyotrophic lateral sclerosis), we find that all five consonant $d'$ features correlate significantly with clinical severity (random-effects meta-analysis $\rho = -0.50$ to $-0.56$, $p < 2 \times 10^{-4}$; pooled Spearman $\rho = -0.47$ to $-0.55$ with bootstrap 95\% CIs not crossing zero), with the effect replicating within individual corpora, surviving FDR correction, and remaining robust to leave-one-corpus-out removal and alignment quality controls. Nasality $d'$ decreases monotonically from control to severe in 6 of 7 severity-graded corpora. Mann-Whitney U tests confirm that all 12 features distinguish controls from severely dysarthric speakers ($p < 0.001$). The method requires no dysarthric training data and applies to any language with an existing MFA acoustic model (currently 29 languages) or a model trained from healthy speech alone. It produces clinically interpretable per-feature profiles. We release the full pipeline and phone feature configurations for six languages to support replication and clinical adoption.

\section{Introduction}
Dysarthria, a group of motor speech disorders caused by neurological damage to the speech production mechanism, affects an estimated 170 per 100,000 in the United Kingdom \cite{ref1} and is a primary symptom of conditions including Parkinson's disease (PD), amyotrophic lateral sclerosis (ALS), cerebral palsy (CP), and stroke. Clinical severity assessment relies on perceptual judgement by trained speech-language pathologists, which is subjective, time-consuming, and unavailable in many care settings. Automated severity classification has therefore attracted growing interest, with recent systems achieving 70\% accuracy on English dysarthric speech using wav2vec2 embeddings and triplet loss \cite{ref2} or 67\% macro F1 across languages using phoneme error metrics \cite{ref3}.

In clinical practice, a speech-language pathologist assesses dysarthria through systematic perceptual evaluation: listening for breathiness in phonation, imprecise consonant articulation, hypernasality from velopharyngeal incompetence, and reduced vowel distinctiveness. These observations map directly onto articulatory subsystems: the larynx (voicing), the velum (nasality), the tongue and lips (manner and place), and the vowel space. The clinician's challenge is that these assessments are subjective, require specialist training, and cannot be performed remotely or continuously. A patient with ALS whose nasality begins to deteriorate, signaling early bulbar involvement, may not be seen by a specialist until weeks after the change began. What is needed is a tool that mirrors the clinician's subsystem-level assessment but operates automatically, objectively, and at any point of care.

However, existing automated approaches share two fundamental limitations. First, they require labelled dysarthric speech data for training, which is scarce for most languages and aetiologies. Second, they produce opaque severity scores that offer clinicians no insight into which articulatory subsystems are degrading, information that is critical for treatment planning and disease monitoring.

Recent work by Choi et al.\ \cite{ref4} demonstrated that self-supervised speech models such as HuBERT \cite{ref5} encode phonological features in linearly separable, near-orthogonal subspaces. Previous-phone, current-phone, and next-phone identities can be decoded with high accuracy using simple linear probes, and these representations are structured along phonologically meaningful dimensions (nasality, voicing, manner of articulation). This finding suggests that the quality of phonological encoding in HuBERT may degrade in a clinically meaningful way when applied to dysarthric speech.

\begin{figure}[H]
\centering
\includegraphics[width=0.95\textwidth]{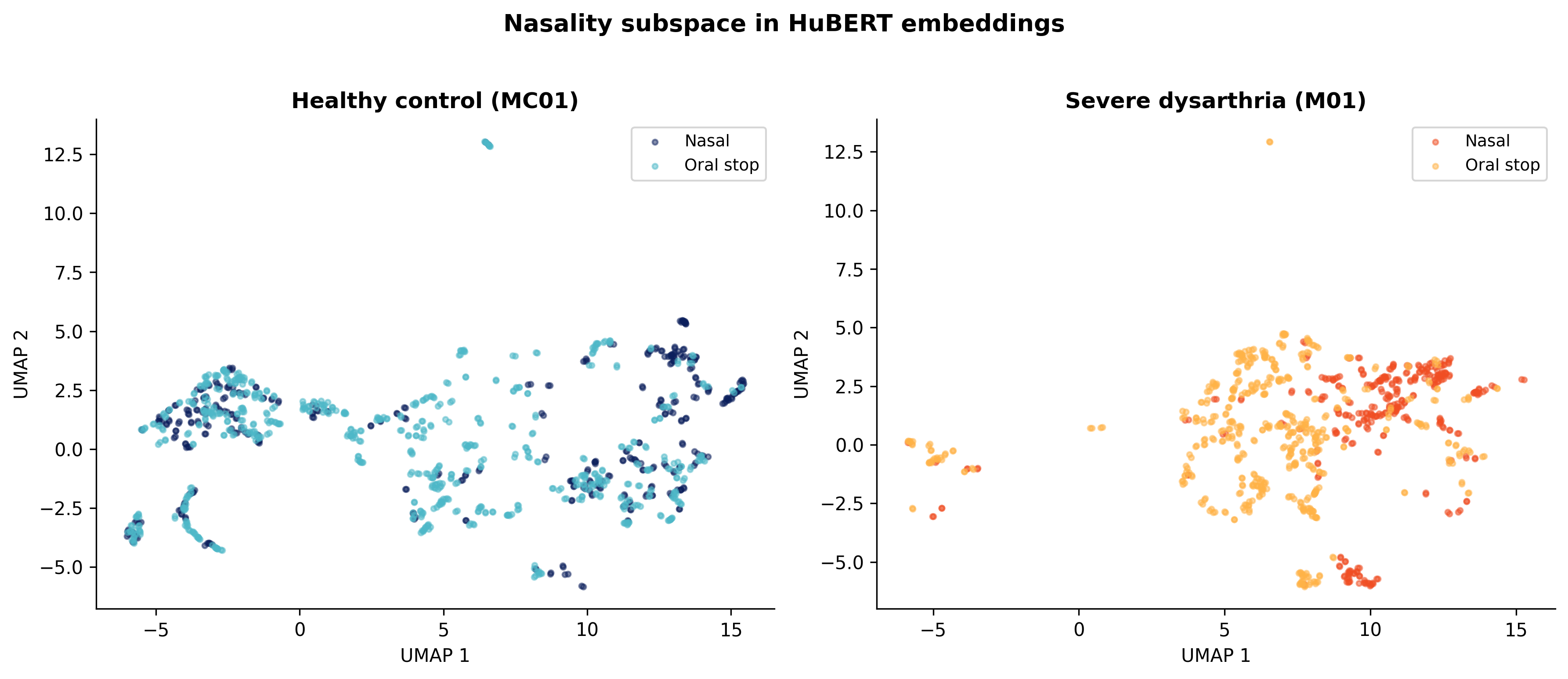}
\caption{Phonological subspace collapse in dysarthric speech. UMAP projection of phone-level HuBERT embeddings for one healthy control speaker (MC01, left) and one speaker with severe dysarthria (M01, right) from the TORGO corpus. Points represent individual phone tokens coloured by class: nasal consonants (m, n, ng) vs.\ oral stops (p, b, t, d, k, g). In the healthy speaker, the two classes form tight, well-separated clusters ($d' = 4.2$). In the severe speaker, the class centroids remain distinguishable but the within-class distributions are wider and overlap substantially ($d' = 1.1$), reflecting degraded articulatory contrast rather than complete phonological merger. This reduction in separability --- quantified as $d'$ --- is the basis of the severity metric developed in this paper. Note that $d'$ is computed along the optimal discriminant direction in 768-dimensional embedding space, not in the 2D UMAP projection shown here.}
\label{fig:fig1}
\end{figure}

We position this method as a research biomarker and clinical screening tool, not as a replacement for expert perceptual evaluation. Its primary use cases are: (i) automated longitudinal monitoring of speech degradation in neurodegenerative disease, (ii) large-scale screening in telehealth settings where specialist access is limited, and (iii) providing interpretable phonological profiles to supplement existing severity classification systems.

\begin{figure}[H]
\centering
\includegraphics[width=0.95\textwidth]{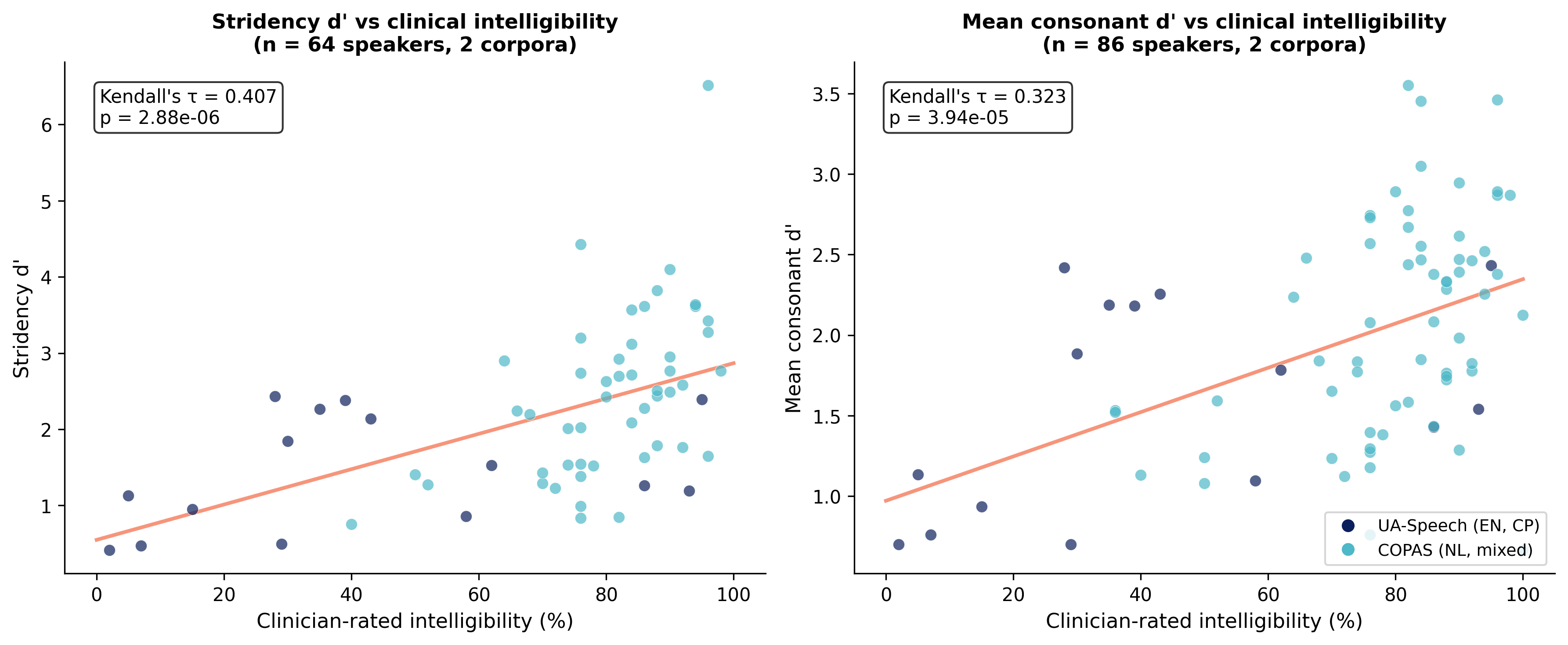}
\caption{Clinical validation: $d'$ correlates with clinician-rated intelligibility across languages. Left: stridency $d'$ vs intelligibility (Kendall's $\tau = 0.407$, $p = 2.9 \times 10^{-6}$, $n = 64$). Right: mean consonant $d'$ vs intelligibility ($\tau = 0.323$, $p = 3.9 \times 10^{-5}$, $n = 86$). Data from UA-Speech (English, cerebral palsy) and COPAS (Dutch, mixed aetiologies). Trendlines show positive linear association.}
\label{fig:fig2}
\end{figure}

We test this hypothesis systematically. Our contributions are:
\begin{itemize}
\item A training-free severity assessment method that measures the degradation of phonological feature subspaces in frozen HuBERT representations, requiring no labelled dysarthric data and no model adaptation.
\item Cross-lingual validation across 890 speakers, 10 corpora, and 5 languages, demonstrating that an English-trained HuBERT model captures severity-related phonological degradation in Spanish, Dutch, Mandarin, and French speech. The method generalises to new languages given a small set of matched healthy control recordings in the target language; no dysarthric speech data is required in any language.
\item Clinically interpretable phonological profiles -- a 12-dimensional vector per speaker that decomposes severity into specific articulatory deficits (nasality, voicing, stridency, sonorance, manner, vowel features, boundary sharpness, and vowel space area), enabling aetiology-specific characterisation.
\item Comprehensive robustness validation including partial correlations, bootstrap CIs, FDR correction, leave-one-corpus-out sensitivity, random-effects meta-analysis, and alignment quality controls, demonstrating that the severity-degradation relationship is not driven by token count confounding, statistical multiplicity, any single corpus, or MFA alignment errors.
\end{itemize}

\section{Related Work}
\subsection{Supervised Severity Classification}
Supervised approaches dominate the dysarthria severity classification literature. Kadirvelu et al.\ \cite{ref2} proposed SALR, a severity-aware learning method using wav2vec2 representations and triplet loss, achieving 70.48\% speaker-independent accuracy on UA-Speech (English, CP). SALR demonstrated that self-supervised speech representations encode severity-relevant information, but the system requires labelled training data and evaluates on a single corpus and language. Joshy and Rajan \cite{ref6} achieved competitive results on UA-Speech using DNN classifiers over hand-crafted features, while Bhat et al.\ \cite{ref7} combined acoustic and articulatory features for severity estimation on TORGO. DSSCNet \cite{ref8} introduced a deep speech severity classifier using spectrogram-based features with attention pooling. Sapkota et al.\ \cite{ref9} conducted layer-wise probing of self-supervised speech models for severity classification, demonstrating that HuBERT's later layers are most informative for severity (73.53\% on TORGO), but their analysis was restricted to English and did not examine the representational structure underlying the severity signal.

A common limitation of all supervised approaches is their dependence on labelled dysarthric training data, which constrains scalability to new languages and clinical populations. None of these systems provides per-feature interpretability of which articulatory subsystems are degrading.

\subsection{Intelligibility Assessment}
A related but distinct task is intelligibility prediction. Venugopalan et al.\ \cite{ref10} developed SpICE at Google, a 5-point intelligibility classifier trained on 551,000 English utterances from Project Euphonia. SpICE demonstrates that large-scale supervised training achieves strong within-English intelligibility prediction (0.93 correlation with UA-Speech), but the system was designed for English only and does not generalise cross-lingually. Troger et al.\ \cite{ref11} proposed a training-free approach to multilingual intelligibility assessment using word error rates from commercial ASR systems (Google, Microsoft), evaluating on German, Czech, and Spanish. Their method shares our goal of avoiding dysarthric training data, but produces only a single intelligibility score rather than a decomposed articulatory profile. Merler et al.\ \cite{ref12} used Whisper with attention analysis for ALS severity regression, achieving $R^2 = 0.92$ on a clinical cohort. Their attention maps identify specific phonemes contributing to severity predictions, paralleling our per-feature analysis, but the approach requires supervised training on labelled ALS speech data.

\subsection{Cross-Lingual Approaches}
Cross-lingual dysarthria assessment remains underexplored. Yeo and Chung \cite{ref13} were among the first to evaluate severity classification across languages (English, Korean, Tamil). More recently, Yeo et al.\ \cite{ref3} proposed multilingual severity estimation using phoneme error rate (PER), phoneme feature error rate (PFER), and phonological coverage (PhonCov) metrics derived from universal phone recognition. Their PhonCov metric -- which measures the loss of phonemic contrasts in ASR output -- is directly complementary to our $d'$ measure: both quantify the degradation of phonological distinctions, but PhonCov operates on ASR decoder output while our method operates on self-supervised encoder representations. Yeo et al.\ evaluated on English, Spanish, Italian, and Tamil, achieving 67.14\% macro F1. Stumpf et al.\ \cite{ref14} explored multilingual dysarthria classification with self-supervised representations but reported accuracies below 65\% and did not provide per-feature interpretability.

Bae et al.\ \cite{ref15} proposed a contrastive learning approach with pseudo-labelling for severity estimation, training on SAP and LibriSpeech and evaluating cross-domain transfer on NeuroVoz, EasyCall, and EWA-DB (0.761 Spearman rank correlation). Their work demonstrates cross-corpus generalisation but frames severity as a regression task rather than providing interpretable articulatory profiles. Bhat and Strik \cite{ref16} provide a comprehensive survey of speech technology for dysarthric speech recognition and assessment, noting that cross-lingual generalisation and clinical interpretability remain open challenges.

\subsection{SSL Representations and Articulatory Information}
Self-supervised speech models have shown promise for pathological speech analysis. Wav2vec2 embeddings serve as the backbone of SALR \cite{ref2}, and HuBERT representations have been used for dysarthric ASR \cite{ref17}. However, these approaches use SSL representations as input features for downstream supervised classifiers rather than analysing the representational structure directly.

Cho et al.\ \cite{ref18} provided critical evidence that SSL representations encode articulatory information, demonstrating strong correlations ($r = 0.81$) between SSL speech features and electromagnetic articulography (EMA) measurements of vocal tract movements. This finding provides theoretical grounding for our approach: if SSL representations encode articulatory gestures, then measuring the degradation of phonological structure in these representations should reflect articulatory impairment.

Choi et al.\ \cite{ref4} extended this line of work by showing that HuBERT encodes phonetic context in position-dependent orthogonal subspaces, with phonological features (nasality, voicing, etc.) linearly decodable from the hidden states. This finding is the theoretical foundation for our method: if healthy speech produces well-separated phonological subspaces, articulatory impairment should measurably degrade this separation, and the pattern of degradation should be clinically informative.

\subsection{Acoustic Vowel Space and $d'$ in Speech Research}
The acoustic vowel space area (VSA) -- the triangle or quadrilateral formed by corner vowel formant frequencies -- is a well-established clinical measure of articulatory working space \cite{ref19,ref20}. VSA reduction correlates with intelligibility loss in PD \cite{ref19}, ALS \cite{ref21}, and CP \cite{ref22}. The formant centralisation ratio (FCR) \cite{ref19} normalises VSA for cross-speaker comparison. Our vowel triangle metric is conceptually analogous to VSA but computed in HuBERT's 768-dimensional embedding space rather than from F1/F2 formant values. This has both advantages (captures voice quality, spectral texture, and temporal dynamics beyond formant structure) and limitations (the learned representation does not decompose into interpretable acoustic dimensions; see Section 6).

Signal detection theory \cite{ref23,ref24} and its sensitivity index $d'$ have been used extensively in speech perception research to quantify the discriminability of phonological contrasts \cite{ref25}. In speech production, $d'$ has been applied to measure acoustic distinctiveness of vowel categories \cite{ref26}. To our knowledge, we are the first to apply $d'$ to phonological feature directions in a self-supervised embedding space as a measure of articulatory precision.

\subsection{Positioning of Our Contribution}
To our knowledge, no prior work combines (i) training-free operation without labelled dysarthric data, (ii) cross-lingual generalisation across five or more languages, and (iii) per-feature clinical interpretability that decomposes overall severity into specific articulatory subsystem profiles. Supervised classifiers \cite{ref2,ref9,ref15} require labelled data; intelligibility predictors \cite{ref10,ref11} produce single scores without articulatory decomposition; cross-lingual phonological metrics \cite{ref3} depend on ASR output quality and do not provide the subsystem-level resolution of our $d'$-based profiles. The present study addresses this gap by measuring phonological subspace degradation in frozen HuBERT representations, requiring only healthy control speech for calibration and producing a 12-dimensional articulatory profile per speaker.

\section{Materials and methods}
\subsection{Overview}
Our method comprises five stages: (1) phone-level forced alignment of speech recordings, (2) extraction of HuBERT frame-level embeddings averaged over phone intervals, (3) computation of phonological feature directions from healthy control speech, (4) measurement of per-speaker $d'$ scores along each feature direction, and (5) correlation of the resulting 12-dimensional phonological profile with clinical severity labels. Stages 1 through 4 use only healthy control data to establish reference directions; dysarthric speakers are evaluated against these directions without any model training or adaptation.

\subsection{Phone-Level Alignment}
Accurate phone-level time boundaries are a prerequisite for computing phone-specific embeddings. We obtain these through the Montreal Forced Aligner (MFA) \cite{ref27}, which performs Viterbi alignment of audio to a phonetic transcription using a pre-trained acoustic model.

We used Montreal Forced Aligner version 3.3.\ \cite{ref27} with the following pretrained acoustic models: english\_mfa, french\_mfa, spanish\_mfa, mandarin\_mfa (from the MFA model repository), and dutch\_cv, italian\_cv (CommonVoice-trained models). Language-specific pronunciation dictionaries were used for all five MFA pipeline languages. For each utterance with an available orthographic transcription, MFA produces a Praat TextGrid file containing phone-level timestamps in IPA notation.

One exception to the standard MFA pipeline exists. For datasets containing sustained vowel recordings without accompanying text (VOC-ALS sustained /a/, /i/, /u/), each audio file is treated as a single phone segment with the vowel identity determined from the filename, and no forced alignment is performed.

A methodological caveat warrants explicit discussion. MFA alignment quality degrades as speech intelligibility decreases: severely dysarthric speech produces more alignment errors (misplaced phone boundaries, spurious insertions, unaligned segments) than healthy speech. Consequently, the $d'$ metrics conflate two sources of signal: (1) genuine degradation of phonological contrasts due to articulatory impairment, and (2) noisier phone embeddings caused by alignment errors. Both factors correlate with severity, so the composite metric remains a valid proxy for severity. However, the contribution of each factor cannot be cleanly separated without ground-truth phonetic segmentation. We address this limitation further in Section 6.

\subsection{HuBERT Embedding Extraction}
All audio was resampled to 16 kHz mono. No amplitude normalisation, silence trimming, or noise reduction was applied.

We use the facebook/hubert-base-ls960 checkpoint from HuggingFace \cite{ref5}, a 12-layer self-supervised speech model pre-trained on 960 hours of English LibriSpeech data, extracting the last hidden state (layer 12), which produces 768-dimensional frame-level representations at 50 frames per second (20 ms per frame). The model is used frozen, with no fine-tuning on dysarthric or multilingual data.

To build intuition for what follows: HuBERT learns, without any labels, to represent speech sounds as points in a high-dimensional space. In this space, sounds that are phonologically similar cluster together -- all /m/ sounds near each other, all /p/ sounds near each other -- and sounds that differ along a phonological dimension (such as nasality) separate along a consistent direction. For a healthy speaker, the boundary between nasal and oral consonants in this space is sharp and well-defined: projecting their phone embeddings onto the nasality direction produces two clearly separated distributions. We quantify this separation using $d'$, a signal detection measure widely used in auditory psychophysics \cite{ref24}. A high $d'$ means the model's representation cleanly distinguishes the two categories; a low $d'$ means they blur together. Our central claim is that dysarthria causes this blurring -- the phonological structure that HuBERT learned from healthy English speech degrades when applied to dysarthric speech, and the degree of degradation correlates with clinical severity.

For each utterance, we extract the final hidden-layer activations across all frames. For each phone interval identified by MFA (Section 3.2), we compute the phone embedding as the mean of all HuBERT frames falling within the interval boundaries, where the set of HuBERT frame indices includes those whose centre falls within the phone interval $[t_{\text{start}}, t_{\text{end}})$, and $h_f$ is the 768-dimensional hidden state at frame $f$. Phone intervals shorter than one frame (20 ms) are extended to include the temporally nearest frame.

\subsection{Feature Direction Computation}
For each language, we compute phonological feature directions from the healthy control speakers within that language. This ensures that the reference directions reflect language-specific phonological realisations (e.g., French nasalised vowels, Mandarin retroflex consonants) rather than imposing a single English-centric template. Note that lexical tone is not modelled in the current feature set; the present analysis is restricted to segmental phonological contrasts.

Each phonological feature $k$ is defined by a binary partition of IPA phones into positive ($P_k$) and negative ($N_k$) classes. We define nine $d'$ features across two categories (five consonant, four vowel), plus three structural metrics (boundary sharpness, cross-position cosine similarity, and vowel triangle area), yielding a 12-dimensional phonological profile per speaker.

\begin{table}[H]
\centering
\footnotesize
\caption*{\textbf{Consonant features (5).}}
\begin{tabular}{llll}
\toprule
Feature & Positive class ($P_k$) & Negative class ($N_k$) & Articulatory basis \\
\midrule
Nasality & /m, n, ng/ & /p, b, t, d, k, g/ & Velopharyngeal port opening \\
Voicing & /b, d, g, v, z, \ldots/ & /p, t, k, f, s, \ldots/ & Vocal fold vibration \\
Stridency & /s, z, sh, zh, f, v, \ldots/ & /p, t, k, b, d, g, m, n, l, r/ & Turbulent airflow at constriction \\
Sonorance & /m, n, ng, l, r, j, w/ & /p, b, t, d, k, g, f, v, s, z, \ldots/ & Periodic voicing without obstruction \\
Manner & /p, b, t, d, k, g, m, n, ng/ & /f, v, s, z, sh, zh, \ldots/ & Complete vs.\ partial oral closure \\
\bottomrule
\end{tabular}
\end{table}

\begin{table}[H]
\centering
\footnotesize
\caption*{\textbf{Vowel features (4).}}
\begin{tabular}{llll}
\toprule
Feature & Positive class & Negative class & Articulatory basis \\
\midrule
High & /i, I, u, U/ & /A, ae, E, O, @, V/ & Tongue height (close) \\
Low & /A, ae/ & /i, I, u, U, @, E/ & Tongue height (open) \\
Back & /u, U, O, A, V/ & /i, I, E, ae/ & Tongue backness \\
Round & /u, U, O/ & /i, I, E, A, ae, @, V/ & Lip rounding \\
\bottomrule
\end{tabular}
\end{table}

MFA outputs language-specific phone symbols that do not correspond to pure IPA. We created per-language mapping files (provided in the supplementary material) that assign each MFA phone symbol to the universal phonological feature classes defined above. These configuration files contain the full phone inventory for each MFA model, ensuring correct classification of language-specific allophones.

The feature direction for feature $k$ is computed as the L2-normalised difference of means between positive and negative class embeddings from healthy controls, where $\bar{m}_{p,k}$ is the mean embedding over all phone tokens belonging to the positive class of feature $k$, pooled across all healthy control speakers for the target language. A minimum of 5 phone tokens per class is required for a direction to be considered valid.

\subsection{Per-Speaker Severity Metrics}
For each speaker, we compute 12 metrics that together form a phonological profile vector. These metrics fall into three categories.

\subsubsection{Consonant $d'$ scores (5 metrics)}
For each consonant feature $k \in \{$nasality, voicing, stridency, sonorance, manner$\}$, we project the speaker's phone embeddings onto the corresponding feature direction $d_k$ and compute $d'$ \cite{ref23}, where $\mu_{p,k}$ and $\mu_{n,k}$ are the mean projections of the speaker's positive and negative class tokens onto $d_k$, and the pooled standard deviation is computed across both classes. Higher $d'$ indicates better preservation of the phonological contrast.

\subsubsection{Vowel $d'$ scores (4 metrics)}
Identically computed for vowel features $k \in \{$high, low, back, round$\}$, using the speaker's vowel phone embeddings projected onto the corresponding vowel feature direction.

\subsubsection{Structural metrics (3 metrics)}
\textbf{Boundary sharpness.} For each consecutive phone pair within an utterance, we compute the cosine similarity of their embeddings. The speaker's boundary sharpness is the mean cosine similarity across all phone transitions. Lower values indicate sharper (more distinct) phone transitions; higher values indicate blurred articulatory boundaries.

\textbf{Cross-position cosine similarity.} For each phone at position $i$ in an utterance (excluding initial and final positions), we compute the cosine similarity between its embedding and the embeddings at positions $i-1$ and $i+1$. This captures the degree to which HuBERT's position-dependent subspace structure \cite{ref4} is preserved.

\textbf{Vowel triangle area.} We compute the mean embedding centroid for each of the three corner vowels (/i/, /a/, /u/) and measure the area of the triangle formed by these centroids in HuBERT embedding space using Heron's formula, where $a, b, c$ are the Euclidean distances between vowel centroids and the semi-perimeter is $s = (a+b+c)/2$. A minimum of 3 tokens per corner vowel is required. This metric is analogous to the acoustic vowel space area \cite{ref19} but computed in the learned representation space rather than from formant frequencies.

\subsection{Severity Mapping}
Datasets in this study use heterogeneous severity labelling schemes. To enable cross-corpus comparison, we adopt the intelligibility-based severity thresholds proposed by Stipancic et al.\ \cite{ref29}:

\begin{table}[H]
\centering
\footnotesize
\caption*{\textbf{Intelligibility-based severity thresholds} (Stipancic et al.\ \cite{ref29}).}
\begin{tabular}{ll}
\toprule
Intelligibility & Severity label \\
\midrule
$> 94\%$ & Control \\
85--94\% & Mild \\
70--84\% & Moderate \\
$< 70\%$ & Severe \\
\bottomrule
\end{tabular}
\end{table}

The Stipancic et al.\ \cite{ref29} framework includes a ``profound'' category ($< 50\%$ intelligibility); in the present study, profound and severe are merged into a single ``severe'' category due to limited sample sizes. These thresholds are applied to all datasets providing intelligibility percentages (COPAS, MDSC, UA-Speech). For datasets with pre-assigned categorical severity labels (SAP, TORGO, PC-GITA, Neurovoz), the original labels are retained. Healthy control speakers from dedicated control corpora (LibriSpeech, UA-Speech controls) are labelled as Control.

\subsection{Datasets}
The full experimental pipeline spans 10 corpora across 5 languages and 3 primary aetiologies, totalling 890 speakers. Of these, 23 COPAS speakers yielded fewer than 25 aligned phone tokens due to short recordings or alignment failures and were excluded from analyses requiring consonant $d'$ computation, leaving 867 speakers in the feature-level analyses. Table~\ref{tab:t1} summarises the speaker counts by severity level.

Throughout this study, the unit of analysis is the individual speaker. Each speaker contributes a single set of 12 metrics, computed from all available aligned utterances. No corpus includes multiple recording sessions for the same speaker in the severity evaluation analysis; the two longitudinal speakers in the proof-of-concept (Section results) are analysed separately and excluded from the main pooled analysis.

\begin{table}[H]
\centering
\footnotesize
\caption{Datasets used in the full analysis pipeline. All 12 metrics computed via MFA-aligned speech except where noted. * SAP has no matched controls; 150 LibriSpeech speakers serve as the English healthy control reference for feature direction computation. ** UA-Speech controls from the UASPEECH\_control partition (13 speakers). *** One speaker recorded at both moderate and severe stages. **** Speakers serving as healthy controls contribute to both direction estimation and severity evaluation (as the control group).}
\label{tab:t1}
\resizebox{\textwidth}{!}{%
\begin{tabular}{lllrrrrrl}
\toprule
Corpus & Language & Aetiology & Control & Mild & Moderate & Severe & Total & Source \\
\midrule
SAP & English & PD, ALS, CP, DS, Stroke & 0* & 159 & 24 & 5 & 188 & Millet et al.\ \cite{ref30} Zheng et al.\ \cite{ref31} \\
COPAS & Dutch & Mixed pathology & 44 & 74 & 77 & 23 & 218 & Martens et al.\ \cite{ref32} \\
TORGO & English & CP, ALS & 7 & 2 & 2 & 4 & 15 & Rudzicz et al.\ \cite{ref33} \\
UA-Speech & English & CP & 13** & 5 & 0 & 10 & 28 & Kim et al.\ \cite{ref34} \\
Neurovoz & Spanish & PD & 58 & 43 & 10 & 0 & 111 & Moro-Velazquez et al.\ \cite{ref35} \\
YouTube\_French & French & ALS & 20 & 1 & 3 & 1*** & 25*** & Collected by authors \\
MDSC & Mandarin & CP & 25 & 11 & 6 & 14 & 56 & Jin et al.\ \cite{ref36} \\
PC-GITA & Spanish & PD & 50 & 30 & 16 & 4 & 100 & Orozco-Arroyave et al.\ \cite{ref37} \\
LibriSpeech & English & HC only & 150 & -- & -- & -- & 150 & Panayotov et al.\ \cite{ref38} \\
Total (severity eval) & 5 langs & & 367 & 325 & 138 & 61 & 891 & All speakers with severity labels \\
\bottomrule
\end{tabular}%
}
\end{table}

Speakers were excluded from individual feature analyses when fewer than 5 phone tokens were available per phonological class. The number of speakers contributing to each feature ($n$) is reported in all tables.

COPAS severity labels are derived from original TSV intelligibility scores via Stipancic et al.\ \cite{ref29} thresholds for all 227 speakers, including 106 speakers with non-standard pathologies (laryngectomy, cleft palate, voice disorder) that were previously unlabelled.

\begin{table}[H]
\centering
\footnotesize
\caption{Additional datasets for vowel triangle analysis only (sustained vowel recordings, no MFA required).}
\label{tab:t2}
\begin{tabular}{lllrrll}
\toprule
Corpus & Language & Aetiology & Control & Dysarthric & Method & Source \\
\midrule
VOC-ALS & Italian & ALS & 72 & 81 & Sustained /a/, /i/, /u/ & Mulfari et al.\ \cite{ref39} \\
PC-GITA & Spanish & PD & 50 & 50 & Sustained /a/, /i/, /u/ & Orozco-Arroyave et al.\ \cite{ref37} \\
\bottomrule
\end{tabular}
\end{table}

\subsection{Statistical Analysis}
We employ the following statistical tests:
\begin{itemize}
\item Spearman rank correlation ($\rho$) between each of the 12 phonological features and ordinal severity (coded as control=0, mild=1, moderate=2, severe=3), computed both pooled across all corpora and within each corpus.
\item Bootstrap 95\% confidence intervals (1,000 iterations with replacement) for all pooled Spearman correlations, providing non-parametric uncertainty estimates.
\item Benjamini-Hochberg FDR correction across 62 total statistical tests (10 overall features + 52 within-corpus tests) at $q = 0.05$, controlling the expected false discovery rate.
\item Partial Spearman correlation controlling for phone token count ($n_{\text{phones}}$) to assess whether token count confounds the severity-$d'$ relationship (Section 5.9.1).
\item Random-effects meta-analysis (DerSimonian-Laird \cite{ref40}) pooling within-corpus correlation estimates with inverse-variance weighting, providing corpus-aware inference that accounts for between-corpus heterogeneity. This serves as the primary inferential frame for the main severity correlations.
\item Mann-Whitney U test for pairwise group comparisons (control vs.\ severe).
\item Logistic regression and leave-one-speaker-out (LOSO) cross-validation for aetiology discrimination from the 12-metric profile.
\item Leave-one-corpus-out sensitivity analysis to assess whether any single corpus drives the pooled correlations.
\end{itemize}

\begin{figure}[H]
\centering
\includegraphics[width=0.95\textwidth]{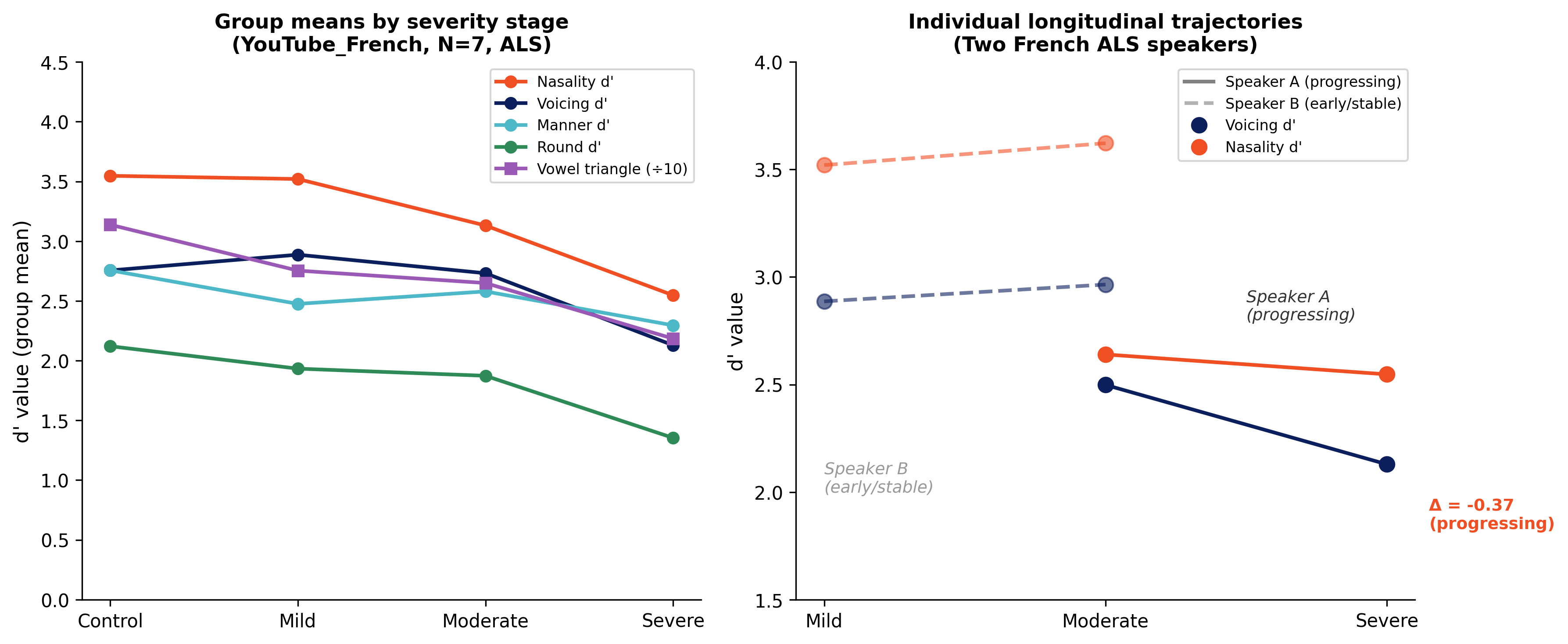}
\caption{Left panel: group means for all reported features across control/mild/moderate/severe stages (Table~\ref{tab:t3}). Right panel: Speaker A (progressing, voicing $d'$ drops 0.37 between moderate and severe) vs.\ Speaker B (stable in early disease). Data from Table~\ref{tab:t3}.}
\label{fig:fig3}
\end{figure}

All $p$-values are two-sided. We report both raw and FDR-corrected $p$-values for the primary analyses (Fig~\ref{fig:fig3}). Of 62 tests, 43 survive FDR correction at $q = 0.05$.

\subsection*{Ethics and Data Governance}
This study performs secondary analysis of previously collected, de-identified speech datasets. All corpora were collected under institutional ethics approvals by their original distributors. As this work involves no direct interaction with human participants and uses only de-identified data, additional institutional review board approval was not required.

The following corpora are publicly available: TORGO \cite{ref32} from the University of Toronto; SAP \cite{ref30} through the University of Illinois under a signed data use agreement; Neurovoz \cite{ref34} on Zenodo with approval from the dataset maintainers; and LibriSpeech \cite{ref38} from OpenSLR under CC BY 4.0.

The following corpora were obtained under data sharing agreements with their respective custodians: COPAS \cite{ref31} from the Dutch Language Institute (IVDNT); UA-Speech \cite{ref33} from the University of Illinois; MDSC \cite{ref35} from AISHELL; PC-GITA \cite{ref36} from the Universidad de Antioquia; and VOC-ALS \cite{ref39} from the original authors.

The YouTube\_French corpus comprises speech from 24 French-speaking individuals. One recording originates from an officially released documentary; informed consent was obtained from one additional participant. The remaining speakers were sourced from publicly posted personal vlogs in which they voluntarily discussed their ALS diagnosis and were not contactable at the time of writing. Speaker identifiers are arbitrary codes with no link to real identities, and no audio is redistributed --- only de-identified aggregate statistics are reported. The extraction of acoustic features from publicly accessible video content for scientific research is permitted under EU Directive 2019/790 Article 3 (text and data mining exception for research organisations) and the processing of de-identified aggregate statistics is consistent with GDPR Article 89 safeguards for scientific research.

\textbf{Data governance.} The CANDOR research programme operates under a patient-centred data governance framework aligned with GDPR principles. All datasets are accessed under the terms specified by their original custodians, and no raw audio is redistributed or stored beyond the scope of the original data sharing agreements. The analysis pipeline processes audio into de-identified aggregate statistical features ($d'$ scores and structural metrics); no speaker-identifiable information is retained in the output. The complete analysis code and phone feature configurations are released as open source, enabling independent replication without requiring access to the underlying audio. This separation of method from data ensures that the research contribution is fully reproducible while respecting participant privacy and data ownership as defined by each corpus's original consent framework.

\subsection*{Reproducibility}
An open-source code repository containing the analysis pipeline, phone feature configuration files for six languages, and all analysis scripts is available for review at \url{https://github.com/Scott-Morgan-Foundation/phonological-subspace-severity}. The repository includes exact package versions (Python 3.12, PyTorch 2.10, HuggingFace Transformers 4.57, Montreal Forced Aligner 3.3), deterministic seeds for all bootstrap analyses (seed=42), and the phone-to-feature mapping files that are central to the method. Raw audio data must be obtained from each corpus's original distributor under their respective data use agreements. The repository is archived at Zenodo (\url{https://doi.org/10.5281/zenodo.19369183}) for permanent access.

\section{Proof-of-Concept}
To validate the approach prior to the full multi-corpus experiment, we conducted a proof-of-concept analysis on 7 French-speaking participants: 3 healthy controls and 4 speakers with ALS recorded at varying severity levels (1 mild, 2 moderate, 1 severe; the moderate and severe recordings include longitudinal data from two individuals). Connected speech samples were aligned using the French MFA acoustic model.

\begin{table}[H]
\centering
\footnotesize
\caption{Proof-of-concept results: severity group means for selected phonological features (YouTube\_French, $N=7$ speakers, ALS aetiology).}
\label{tab:t3}
\begin{tabular}{lrrrrr}
\toprule
Metric & Control & Mild & Moderate & Severe & Ctrl-to-Sev change \\
\midrule
Nasality $d'$ & 3.55 & 3.57 & 3.15 & 2.55 & $-28.2\%$ \\
Voicing $d'$ & 2.75 & 2.90 & 2.74 & 2.14 & $-22.2\%$ \\
Manner $d'$ & 2.82 & 2.51 & 2.59 & 2.29 & $-18.8\%$ \\
Round $d'$ & 2.12 & -- & 1.66 & 1.35 & $-36.2\%$ \\
Low $d'$ & 1.86 & -- & 1.82 & 1.87 & $+0.8\%$ \\
Boundary sharpness & 0.354 & 0.424 & 0.487 & 0.524 & $+48.0\%$ \\
Cross-position cosim & 0.165 & 0.173 & 0.176 & 0.195 & $+18.2\%$ \\
Vowel triangle area & 31.38 & -- & 25.65 & 21.85 & $-30.4\%$ \\
\bottomrule
\end{tabular}
\end{table}

Eleven of 12 metrics show monotonic degradation from control to severe. The sole exception is low $d'$ ($+0.8\%$), which measures the distinction between low vowels (/a/, /ae/) and non-low vowels. This is consistent with the clinical observation that /a/ is the last vowel to deteriorate in ALS, as it requires the least precise articulatory positioning \cite{ref41}.

Longitudinal analysis of one ALS speaker (Speaker A, recorded at moderate and severe stages) suggests voicing $d'$ as a candidate progression marker (2.50 to 2.14, $\Delta = -0.37$), while a second speaker (Speaker B, mild to moderate) shows stable $d'$, consistent with early ALS where phonological-level degradation has not yet manifested.

These results are preliminary ($N=7$), but they motivated the full-scale analysis described in the following section.

\section{Results}
\subsection{Within-Corpus Severity Correlations}

\begin{figure}[H]
\centering
\includegraphics[width=0.95\textwidth]{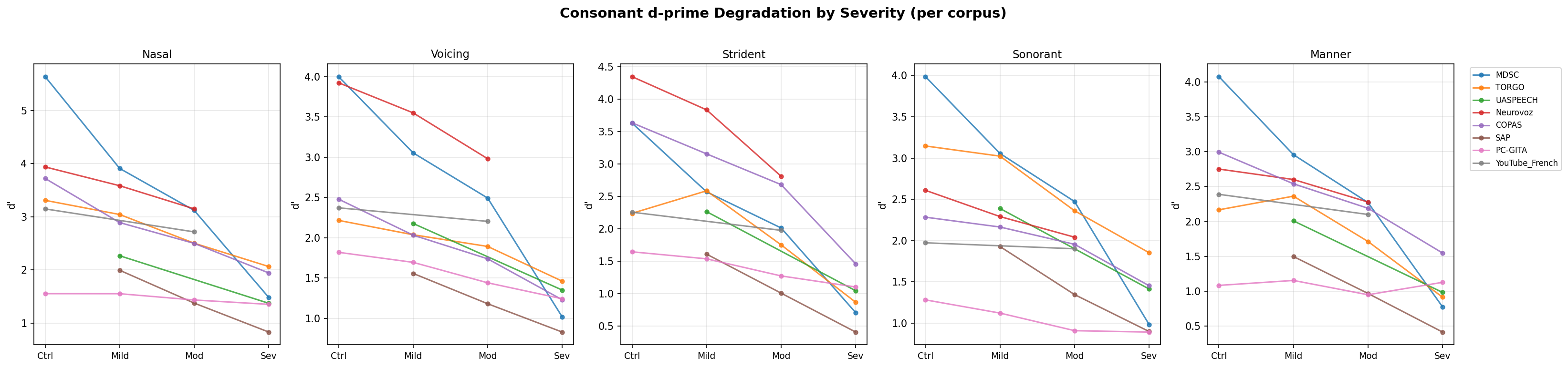}
\caption{Consonant $d'$ degradation by severity across corpora. Each panel displays one of the five consonant $d'$ features (nasality, sonorant, voicing, strident, manner) with speaker-level means plotted across severity groups (control, mild, moderate, severe) for each corpus independently. Lines connect severity group means within each corpus. All five features show a consistent downward trend from control to severe, with the steepest declines in MDSC (Mandarin, CP) and UA-Speech (English, CP). Corpora with fewer than three severity levels are omitted from individual panels.}
\label{fig:fig4}
\end{figure}

Table~\ref{tab:t4} shows that the severity-degradation relationship holds within individual corpora, ruling out a pooling artifact driven by between-corpus differences in language, severity distribution, or recording conditions. Table~\ref{tab:t4} reports within-corpus Spearman correlations for the five consonant features.

\begin{table}[H]
\centering
\footnotesize
\caption{Within-corpus severity correlations demonstrate the effect is not a pooling artifact. Spearman $\rho$ for consonant $d'$ features vs.\ ordinal severity. Only corpora with at least 3 severity levels shown. * $p < 0.05$, ** $p < 0.01$, *** $p < 0.001$.}
\label{tab:t4}
\begin{tabular}{lrrrrr}
\toprule
Corpus (lang, aetiology) & Nasality & Sonorant & Voicing & Strident & Manner \\
\midrule
MDSC (zh, CP) & $-0.923$*** & $-0.910$*** & $-0.879$*** & $-0.901$*** & $-0.912$*** \\
UA-Speech (en, CP) & $-0.786$*** & $-0.733$*** & $-0.655$*** & $-0.726$*** & $-0.691$*** \\
TORGO (en, CP/ALS) & $-0.735$*** & $-0.681$*** & $-0.659$*** & $-0.602$** & $-0.687$*** \\
Neurovoz (es, PD) & $-0.514$*** & $-0.453$*** & $-0.349$*** & $-0.412$*** & $-0.387$*** \\
COPAS (nl, mixed) & $-0.478$*** & $-0.411$*** & $-0.325$*** & $-0.389$*** & $-0.356$*** \\
SAP (en, mixed) & $-0.404$*** & $-0.394$*** & $-0.374$*** & $-0.381$*** & $-0.376$*** \\
\bottomrule
\end{tabular}
\end{table}

The within-corpus correlations are consistently negative across all corpora and all features. The strongest within-corpus effects appear in MDSC (Mandarin, CP; $\rho = -0.88$ to $-0.92$), UA-Speech (English, CP; $\rho = -0.66$ to $-0.79$), and TORGO (English, CP/ALS; $\rho = -0.60$ to $-0.74$). The correlation is weaker but still significant in Neurovoz, COPAS, and SAP. This confirms that the severity-degradation relationship holds within individual datasets, independent of any cross-corpus confounds.

MDSC contains exclusively CP speakers with wide severity range and relatively uniform recording conditions, providing an ideal testbed for the method. The weaker correlations in SAP and COPAS likely reflect their mixed-aetiology composition, heterogeneous recording conditions, and in the case of SAP, the absence of matched controls (LibriSpeech speakers serve as the healthy reference, introducing a recording environment mismatch).

\subsection{Overall Severity Correlations and Group Discrimination}
\begin{figure}[H]
\centering
\includegraphics[width=0.95\textwidth]{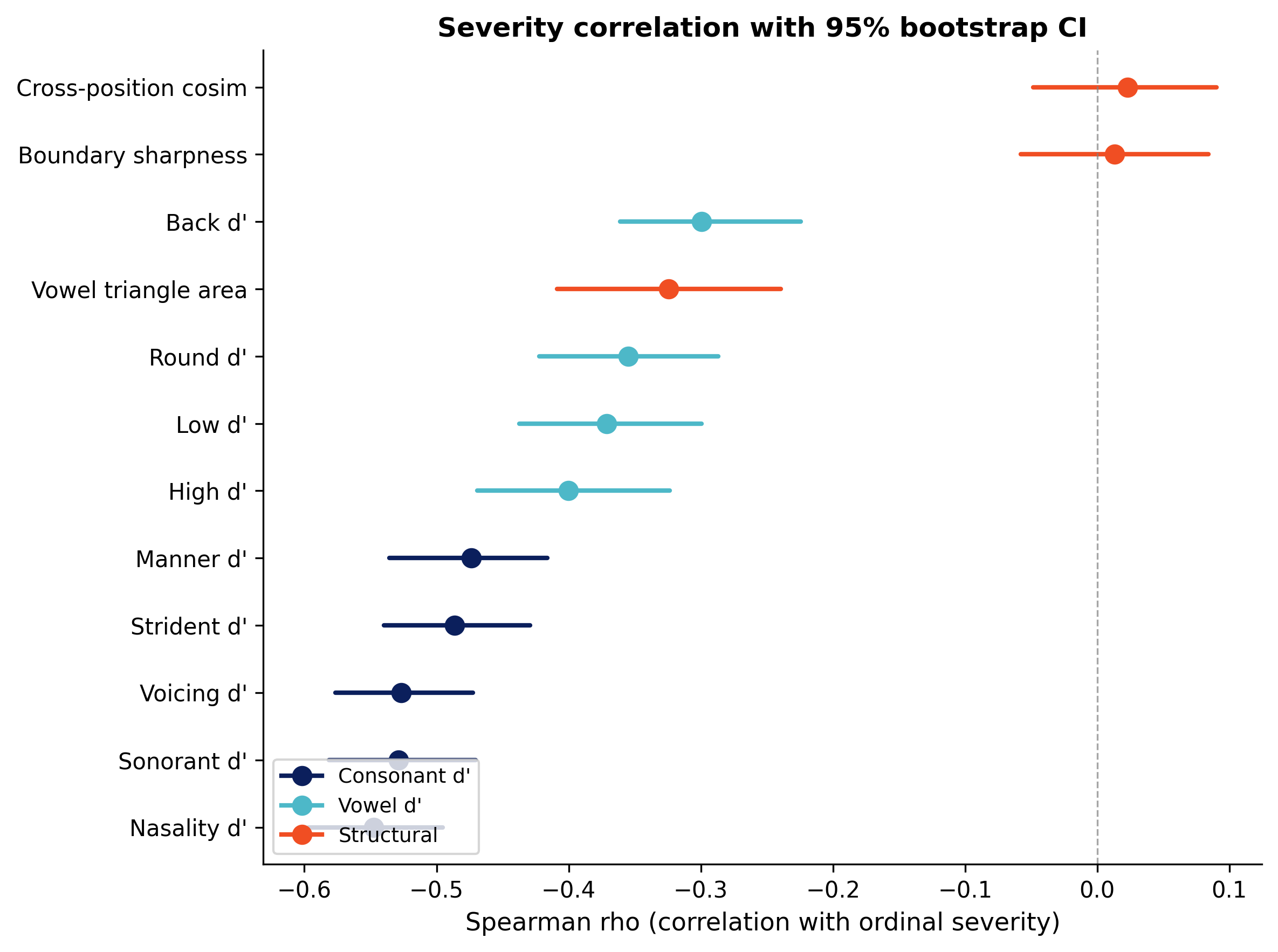}
\caption{Severity correlation forest plot for all 12 phonological features. Spearman $\rho$ between each feature and ordinal severity (control $= 0$, mild $= 1$, moderate $= 2$, severe $= 3$), pooled across all corpora. Horizontal bars show bootstrap 95\% confidence intervals (1,000 iterations). Features are sorted by effect size. Dark navy: consonant $d'$ features; light blue: vowel $d'$ features; orange: structural metrics. All $d'$ features correlate negatively with severity (CIs not crossing zero); boundary sharpness and cross-position cosine similarity show near-zero pooled correlations due to the speech-type confound discussed in the text.}
\label{fig:fig5}
\end{figure}

Fig~\ref{fig:fig5} presents pooled Spearman correlations with bootstrap 95\% confidence intervals for all 12 features, sorted by effect size.

Bootstrap CIs (5,000 iterations, seed=42) for all five consonant features are narrow (0.10--0.12 width) and none cross zero, confirming the stability of the correlation estimates. Cliff's delta effect sizes are uniformly large ($d > 0.63$) for all $d'$ features, with stridency showing the strongest group separation ($d = 0.915$). All 10 overall feature correlations survive Benjamini-Hochberg FDR correction at $q = 0.05$; only boundary sharpness and cross-position cosine similarity fail to survive correction, consistent with their non-significance in the Spearman analysis.

All five consonant $d'$ features correlate negatively with severity ($\rho = -0.47$ to $-0.55$), as shown in Figure~\ref{fig:fig5}, confirming that phonological feature separation in HuBERT space degrades as dysarthria worsens. The four vowel $d'$ features show moderate but significant correlations ($\rho = -0.30$ to $-0.40$). All 12 features significantly distinguish controls from severely dysarthric speakers in the Mann-Whitney U test ($p < 0.001$), including boundary sharpness and cross-position cosine similarity, which do not reach significance in the Spearman analysis --- a finding we examine further in Section 5.4. Stridency $d'$ shows the largest U statistic in the control-vs-severe comparison, suggesting that fricative production is particularly vulnerable to severe articulatory impairment across aetiologies.

Nasality $d'$ emerges as the single strongest correlate of ordinal severity ($\rho = -0.547$, 95\% CI $[-0.597, -0.495]$), followed by sonorant $d'$ ($-0.529$ $[-0.578, -0.468]$) and voicing $d'$ ($-0.527$ $[-0.576, -0.475]$). These pooled correlations are further validated by random-effects meta-analysis across corpora (Section 5.9.5), which serves as the primary corpus-aware inferential frame for this study.

\begin{figure}[H]
\centering
\includegraphics[width=0.95\textwidth]{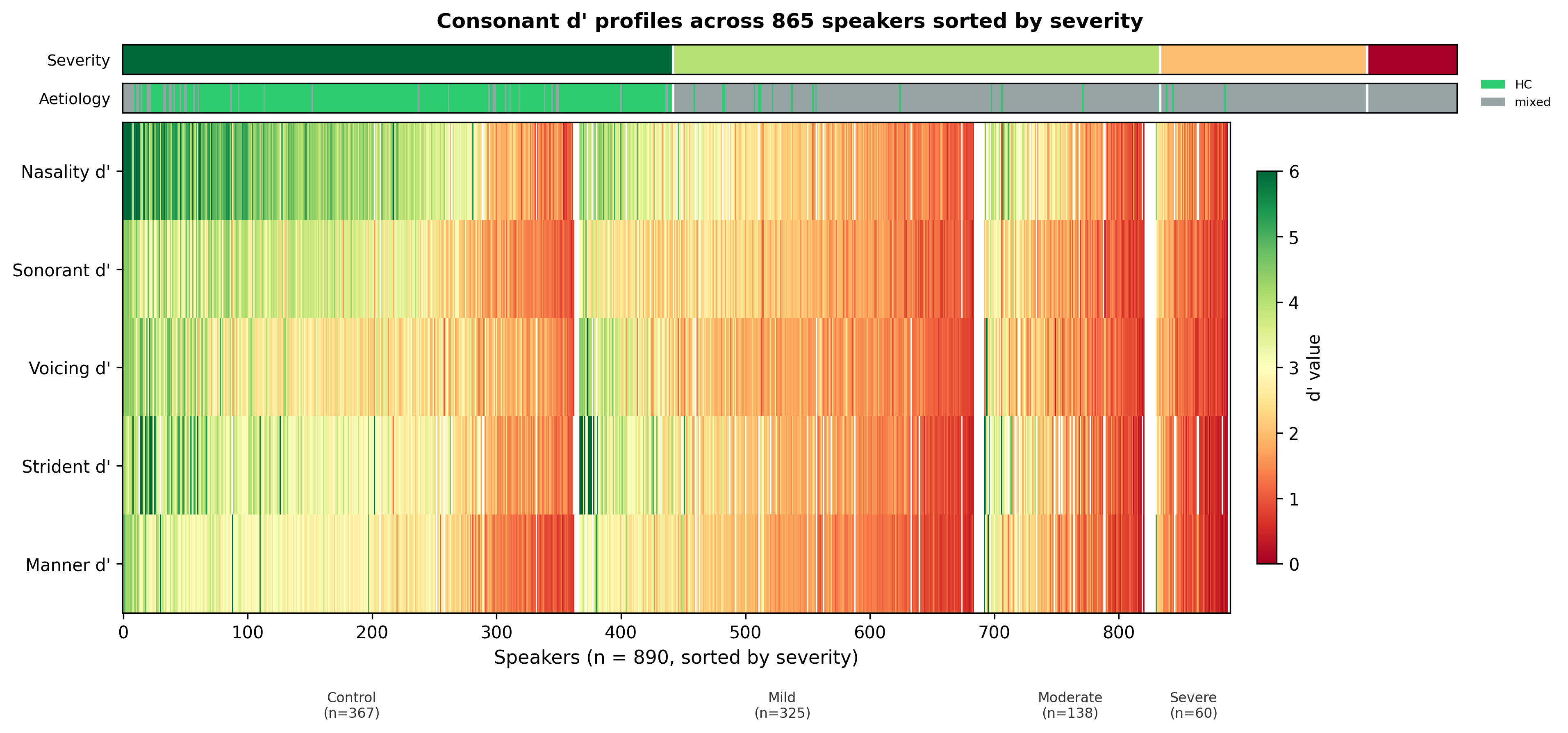}
\caption{Consonant $d'$ profiles across 867 speakers sorted by severity. Each column is one speaker; each row is one consonant $d'$ feature. Speakers are sorted left to right by severity (control, mild, moderate, severe), then by mean $d'$ within each group. Top strips show severity (green to red) and aetiology (colour-coded). The green-to-red gradient across all five features demonstrates that phonological subspace degradation is consistent, gradual, and present across aetiologies and corpora.}
\label{fig:fig6}
\end{figure}

\subsection{Cross-Lingual Consistency}
A key claim of this work is that phonological subspace degradation is language-independent. Table~\ref{tab:t5} shows that consonant $d'$ values decrease monotonically from control to severe across all five evaluation languages, despite HuBERT being pre-trained exclusively on English speech.

\begin{table}[H]
\centering
\footnotesize
\caption{Mean consonant $d'$ by severity and language. Values are means across speakers within each severity-language cell. * Spanish severe values from PC-GITA ($n=4$ severe speakers).}
\label{tab:t5}
\begin{tabular}{llrrrrr}
\toprule
Feature & Severity & English & Spanish & Dutch & Mandarin & French \\
\midrule
Nasality $d'$ & Control & 4.31 & 2.84 & 3.72 & 5.64 & 3.55 \\
 & Severe & 1.57 & 1.35* & 1.94 & 1.48 & 2.55 \\
Strident $d'$ & Control & 2.95 & 3.09 & 2.77 & 3.63 & 2.25 \\
 & Severe & 0.99 & 1.10* & 1.21 & 0.71 & 1.76 \\
Sonorant $d'$ & Control & 3.65 & 2.00 & 1.86 & 3.99 & 1.97 \\
 & Severe & 1.54 & 0.89* & 1.29 & 0.98 & 1.70 \\
\bottomrule
\end{tabular}
\end{table}

The absolute $d'$ values differ substantially across languages (Mandarin controls show higher $d'$s than Spanish controls), which partly reflects language-specific phonological inventories and partly reflects the token count confound discussed in Section 5.6. The direction of the effect is consistent across all languages: $d'$ decreases with increasing severity.

Nasality $d'$ decreases monotonically from control to severe in 6 of 7 corpora that have severity-graded speakers, with the sole exception being a corpus where the severe cell contains fewer than 5 speakers.

\subsection{Boundary Sharpness: A Content-Type-Dependent Effect}
Boundary sharpness and cross-position cosine similarity do not reach significance in the pooled Spearman analysis (Fig~\ref{fig:fig5}), despite showing strong monotonic increases in the proof-of-concept (Table~\ref{tab:t3}). Detailed within-corpus analysis reveals a content-type interaction: boundary sharpness decreases with severity in spontaneous speech but increases in read speech.

\begin{table}[H]
\centering
\footnotesize
\caption{Boundary sharpness vs.\ severity stratified by speech type (Spearman $\rho$). * $p < 0.05$, *** $p < 0.001$, ns = not significant.}
\label{tab:t6a}
\begin{tabular}{lllll}
\toprule
Speech type & $\rho$ & $p$ & $n$ (speakers) & Direction \\
\midrule
Read sentences & $-0.10$* & 0.01 & -- & Slight blurring \\
Spontaneous speech & $-0.12$ ns & -- & 24 & Blurring (expected) \\
Single words & $+0.28$ ns & -- & 28 & Sharpening \\
Mixed (SAP) & $+0.32$*** & 5e-6 & 188 & Sharpening \\
\bottomrule
\end{tabular}
\end{table}

\begin{table}[H]
\centering
\footnotesize
\caption{Boundary sharpness vs.\ severity by corpus (Spearman $\rho$). ** $p < 0.01$, *** $p < 0.001$, ns = not significant.}
\label{tab:t6b}
\begin{tabular}{llll}
\toprule
Corpus & Content type & $\rho$ & Direction \\
\midrule
PC-GITA & Read sentences & $+0.46$*** & Sharpening \\
MDSC & Read words & $+0.47$*** & Sharpening \\
SAP & Mixed tasks & $+0.32$*** & Sharpening \\
COPAS & Read sentences & $+0.18$** & Sharpening \\
YouTube\_French & Spontaneous / connected & $-0.12$ ns & Blurring \\
\bottomrule
\end{tabular}
\end{table}

The task-stratified analysis (Table~\ref{tab:t6a}) reveals that the overall pooled null result masks opposing task-dependent effects: read speech shows a slight decrease in boundary sharpness with severity, while single-word and mixed-task speech shows increased sharpness. However, the per-corpus analysis (Table~\ref{tab:t6b}) demonstrates that this interaction is better characterised as corpus-dependent rather than purely task-dependent: PC-GITA and MDSC (both primarily read speech/words) show the strongest positive correlations, while YouTube\_French (spontaneous speech) shows a non-significant negative trend.

Two competing hypotheses may explain this pattern. First, the compensatory hyperarticulation hypothesis: dysarthric speakers reading aloud may deliberately slow their speech and exaggerate articulatory targets, producing boundaries that are acoustically sharper than those of fluent healthy speakers. Second, the frame stability hypothesis: the slow, paused nature of dysarthric read speech may provide HuBERT with longer stretches of acoustically stable frames within each phone interval, leading to higher boundary contrast as a mathematical consequence of reduced within-phone variability rather than genuine articulatory overshoot. Distinguishing these hypotheses requires concurrent articulatory tracking data (e.g., electromagnetic articulography) alongside acoustic recordings --- data that is not available for the corpora in this study. We present the task-type interaction as an empirical observation and refrain from strong causal claims about the underlying mechanism.

This finding has methodological implications: boundary sharpness is a valid severity metric only within a controlled speech content type and recording condition. Pooling across speech tasks and corpora cancels the effect. Future work should stratify analyses by speech task.

\subsection{Per-Aetiology Phonological Profiles}
\begin{figure}[H]
\centering
\includegraphics[width=0.95\textwidth]{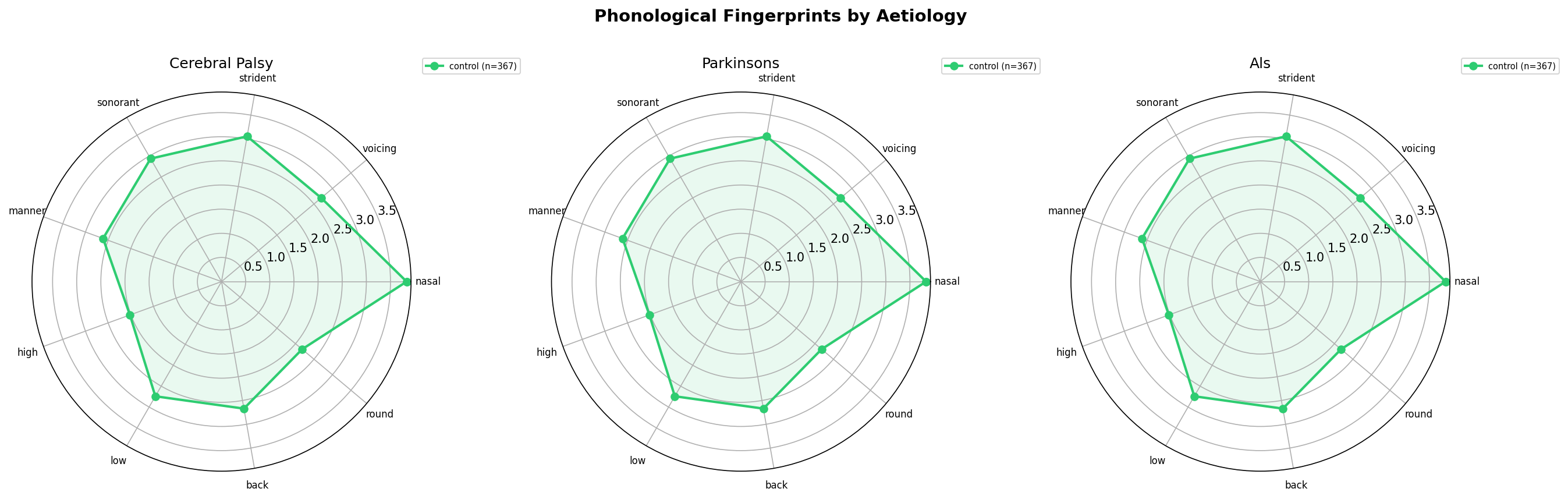}
\caption{Phonological fingerprints by aetiology. Radar plots showing the mean consonant $d'$ profile for cerebral palsy (CP, $n = 86$), Parkinson's disease (PD, $n = 211$), and amyotrophic lateral sclerosis (ALS, $n = 50$) speakers. Each axis represents one consonant $d'$ feature; the green shaded area shows the healthy control reference profile. CP shows the most uniform degradation across all features, while PD and ALS exhibit relatively preserved voicing. These aetiology-specific profiles demonstrate that the method captures clinically meaningful differences in articulatory impairment patterns.}
\label{fig:fig7}
\end{figure}

Table~\ref{tab:t7} presents the Spearman correlations stratified by aetiology. Figure~\ref{fig:fig7} displays the per-aetiology phonological profiles as radar plots.

\begin{table}[H]
\centering
\footnotesize
\caption{Per-aetiology Spearman $\rho$ for consonant $d'$ features vs.\ severity. Only aetiologies with $n \geq 50$ speakers and 3+ severity levels shown. * ALS limited to 50 speakers from SAP and YouTube\_French, with only 5 severity-rated (YouTube\_French); SAP ALS speakers are predominantly mild. ** $p < 0.01$ despite small $n$; strong trend but limited statistical power.}
\label{tab:t7}
\begin{tabular}{lrrr}
\toprule
Feature & CP ($n=86$) & PD ($n=211$) & ALS ($n=50$*) \\
\midrule
Nasality $d'$ & $-0.47$ & $-0.38$ & $-0.365$** \\
Sonorant $d'$ & $-0.42$ & $-0.35$ & $-0.365$** \\
Voicing $d'$ & $-0.35$ & $-0.29$ & $-0.28$ \\
Strident $d'$ & $-0.44$ & $-0.33$ & $-0.317$** \\
Manner $d'$ & $-0.30$ & $-0.31$ & $-0.30$ \\
Boundary sharpness & $+0.27$ & -- & -- \\
Vowel triangle & -- & significant & -- \\
\bottomrule
\end{tabular}
\end{table}

Cerebral palsy shows the strongest correlations across all consonant features ($\rho = -0.30$ to $-0.47$), consistent with the severe articulatory impairment characteristic of CP. Boundary sharpness increases with severity in CP ($\rho = +0.27$), reflecting the read-speech compensation effect from Section 5.4; the CP datasets here (TORGO, UA-Speech, MDSC) consist primarily of read words and sentences.

Parkinson's disease shows moderate, broadly distributed correlations across all features ($\rho = -0.29$ to $-0.38$), with no single dominant degradation. The vowel triangle is significantly correlated with severity, consistent with the well-documented vowel space reduction in PD \cite{ref19,ref20}. The PD profile, moderate, diffuse degradation without a single dominant deficit, is clinically plausible given the global hypokinetic dysarthria characteristic of PD.

ALS shows strong trends (strident $\rho = -0.317$, sonorant $\rho = -0.365$), but statistical power is limited by the small number of severity-rated ALS speakers ($n=5$ from YouTube\_French with severity labels; SAP ALS speakers are predominantly mild). The nasality degradation predicted from clinical literature (velopharyngeal weakness in bulbar ALS) is supported by the data but requires larger ALS cohorts for confirmation.

\subsection{The Token Count Confound}
A critical finding emerged during cross-corpus comparison: $d'$ is positively correlated with the number of phone tokens per speaker, even in healthy controls. Table~\ref{tab:t8} illustrates this.

\begin{table}[H]
\centering
\footnotesize
\caption{Spearman correlation between $d'$ and phone token count ($n_{\text{phones}}$) in healthy control speakers only.}
\label{tab:t8}
\begin{tabular}{lr}
\toprule
Feature & $\rho$ ($d'$ vs $n_{\text{phones}}$, HC only) \\
\midrule
Sonorant $d'$ & $+0.554$ \\
Nasality $d'$ & $+0.405$ \\
Voicing $d'$ & $+0.380$ \\
Strident $d'$ & $+0.350$ \\
\bottomrule
\end{tabular}
\end{table}

\begin{figure}[H]
\centering
\includegraphics[width=0.95\textwidth]{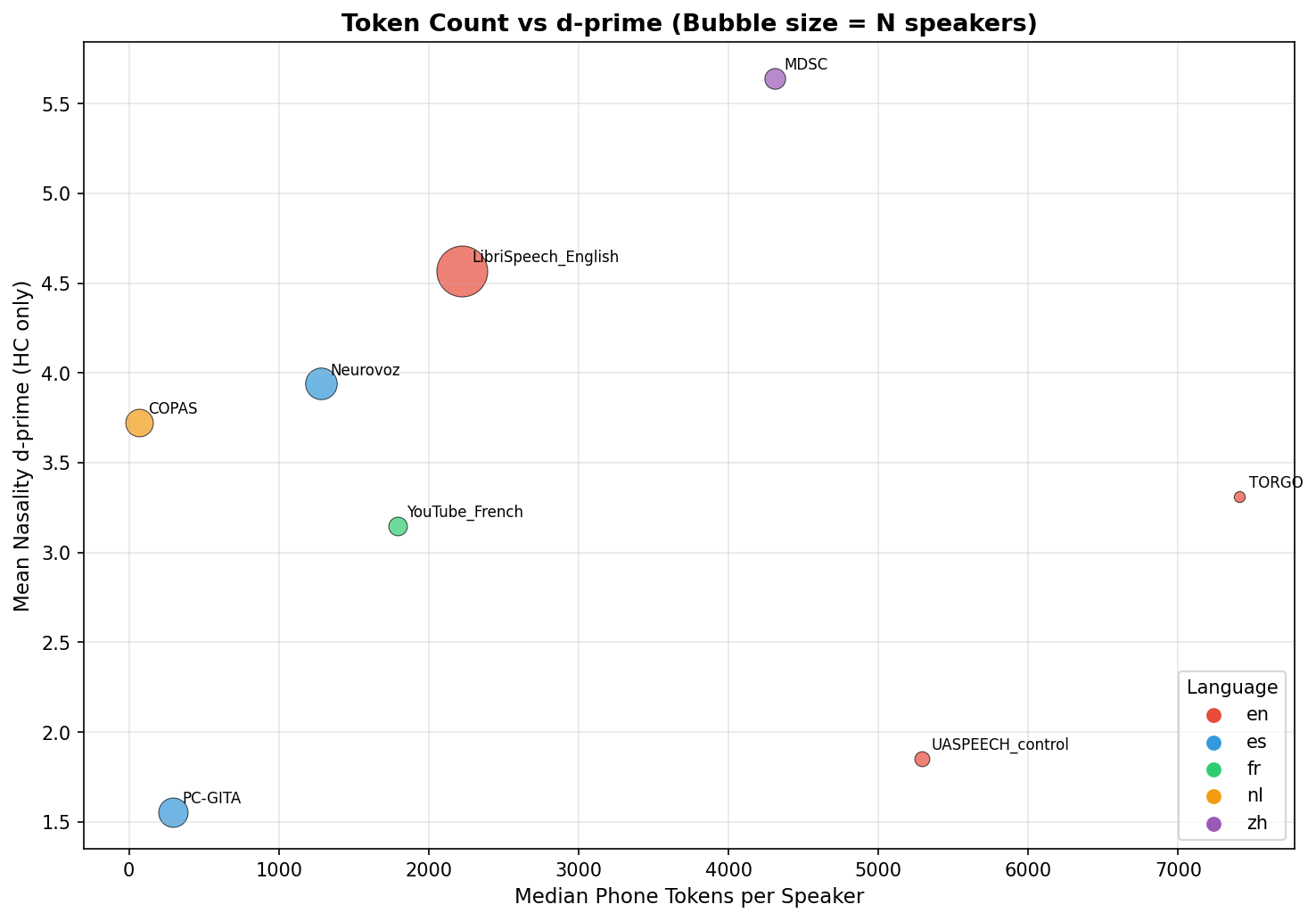}
\caption{Token count versus nasality $d'$ across corpora. Each bubble represents one corpus, positioned by median phone tokens per speaker ($x$-axis) and mean nasality $d'$ in healthy controls only ($y$-axis). Bubble size is proportional to the number of speakers; colour indicates language. The positive association between token count and $d'$ in healthy controls reveals a confound: corpora with shorter utterances yield lower $d'$ estimates regardless of severity, motivating the partial correlation analyses reported in the Token Count Confound section.}
\label{fig:fig8}
\end{figure}

Figure~\ref{fig:fig8} visualises this relationship. This occurs because $d'$ estimation is biased upward with more observations -- the sample means more closely approximate the true population means when $n$ is large, reducing the pooled standard deviation relative to the mean difference. The practical consequence is that absolute $d'$ values are not comparable across corpora with different amounts of speech per speaker. For example, PC-GITA (median 296 phones per speaker) produces $d'$s 2x lower than Neurovoz (median 1,283 phones per speaker), despite both being Spanish PD corpora.

This confound does not invalidate the severity correlations. We confirmed this empirically using partial Spearman correlations that control for $n_{\text{phones}}$ (Section 5.9.1): all consonant $d'$ severity correlations survive with virtually unchanged magnitude (e.g., nasality $\rho = -0.547$ raw vs.\ $-0.548$ partial), and stridency $d'$ actually strengthens from $-0.486$ to $-0.504$ after partialling out token count. Stratification by token-count quartile further confirms that correlations are weakest in the lowest-token speakers (Q1: $\rho = -0.13$, reflecting noisy $d'$ estimates) and not inflated in high-token speakers (Q4: $\rho = -0.37$). The token count confound primarily affects cross-corpus comparison of absolute $d'$ values and explains some of the between-language variation in Table~\ref{tab:t5}. Future work requiring cross-corpus severity comparison (rather than within-corpus ranking) should normalise $d'$ by token count or adopt a fixed-length sampling strategy.

\subsection{Vowel Triangle in HuBERT Space}
The vowel triangle metric produces clear results for connected vowels in ALS but more complex patterns for PD sustained vowels.

\begin{table}[H]
\centering
\footnotesize
\caption{HuBERT vowel triangle area by severity (sustained vowel datasets). * PC-GITA severe cell contains only 4 speakers; the bounce at severe is likely a small-sample artifact. ** Control-to-moderate change.}
\label{tab:t9}
\begin{tabular}{lrrrrr}
\toprule
Corpus (aetiology) & Control & Mild & Moderate & Severe & Ctrl-to-Sev \\
\midrule
VOC-ALS (ALS) & 28.79 & 27.45 & 22.73 & 17.31 & $-39.9\%$ \\
PC-GITA (PD) & 21.21 & 19.54 & 17.17 & 19.12* & $-9.3\%$** \\
\bottomrule
\end{tabular}
\end{table}

VOC-ALS shows clear monotonic degradation of the vowel triangle (40\% reduction from control to severe), consistent with the global articulatory collapse in ALS. PC-GITA shows degradation from control to moderate but a bounce at severe ($n=4$), consistent with either a small-sample artifact or PD-specific hyperarticulation of sustained vowels.

An important caveat: the HuBERT vowel triangle is not equivalent to the classical formant-based vowel space area. HuBERT embeddings encode voice quality, breathiness, tremor, and spectral texture in addition to formant structure. For sustained vowels in PD, vocal pathology (breathiness, tremor) may create more dispersed embeddings, inflating the triangle even as the formant-based VSA shrinks. The HuBERT vowel triangle is therefore most informative for conditions involving articulatory collapse (ALS, severe CP) rather than conditions primarily affecting voice quality (mild-moderate PD).

\subsection{Aetiology Discrimination}
We tested whether the 12-metric phonological profile can distinguish aetiologies using logistic regression with LOSO cross-validation across the three primary aetiologies (PD, CP, ALS).

Overall accuracy: 45.2\% (chance $= 33.3\%$ for 3-class).

\begin{table}[H]
\centering
\footnotesize
\caption*{\textbf{Per-aetiology LOSO recall.}}
\begin{tabular}{ll}
\toprule
Aetiology & Recall \\
\midrule
PD & 53\% \\
CP & 76\% \\
ALS & low (limited by $n$) \\
\bottomrule
\end{tabular}
\end{table}

The most discriminative features were round $d'$ (coefficient 1.08), sonorant $d'$ (0.96), and high $d'$ (0.79). CP is the most distinguishable aetiology (76\% recall), likely because CP produces the most severe and broadly distributed articulatory impairment. PD recall is moderate (53\%), and ALS discrimination is poor due to the limited number of ALS speakers.

At 45.2\%, well above the 33.3\% chance baseline, the result confirms that the phonological profile carries aetiology-discriminative information. The method was designed as a descriptive profiling tool, not an aetiology classifier. The aetiology discrimination result serves primarily to validate that different aetiologies produce different phonological degradation patterns, supporting the clinical interpretability of the profiles.

\subsection*{Screening Threshold Analysis}
\begin{figure}[H]
\centering
\includegraphics[width=0.95\textwidth]{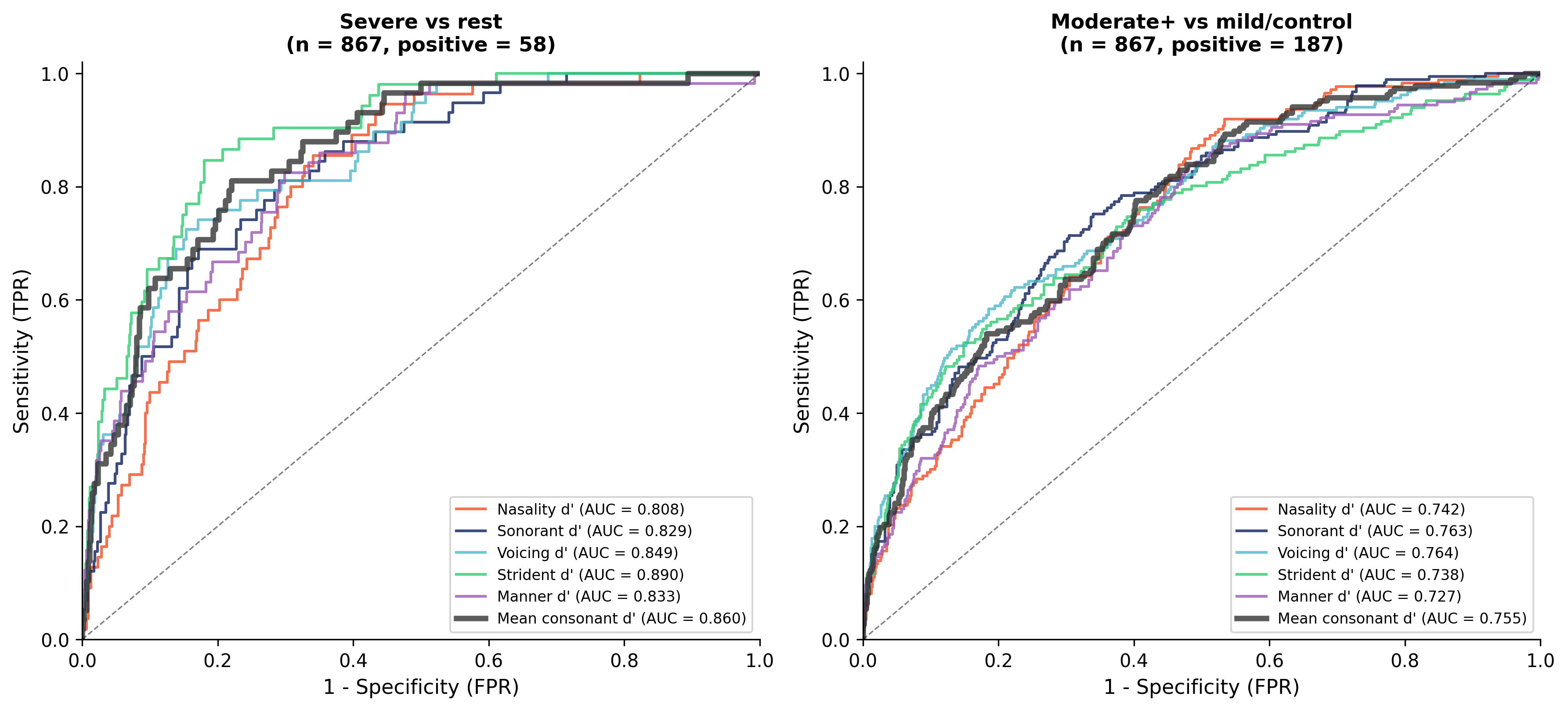}
\caption{ROC curves for binary severity detection. Left: severe vs.\ rest (stridency $d'$ AUC $= 0.890$). Right: moderate-or-worse vs.\ mild/control. All five consonant $d'$ features and their mean are shown. Optimal thresholds determined by Youden's J statistic.}
\label{fig:fig9}
\end{figure}

To assess the method's potential as a clinical screening tool, we computed receiver operating characteristic (ROC) curves for two binary classification tasks: detecting severe dysarthria (severe vs.\ all other speakers) and detecting moderate-or-worse dysarthria (moderate + severe vs.\ mild + control). Fig~\ref{fig:fig9} presents ROC curves for each consonant $d'$ feature and their mean. For detecting severe dysarthria ($n = 58$ severe, 809 other), stridency $d'$ achieves AUC $= 0.890$ with 85\% sensitivity and 82\% specificity at an optimal threshold of $d' = 1.43$. Mean consonant $d'$ achieves AUC $= 0.860$. All individual features exceed AUC $= 0.80$. For the harder task of detecting moderate-or-worse dysarthria ($n = 187$ positive, 680 negative), AUC values range from 0.727 to 0.764, with sonorant and voicing $d'$ performing best. These results suggest that consonant $d'$ features, particularly stridency, could support automated screening for severe speech deterioration in clinical and telehealth settings.

\subsection{Robustness and Sensitivity Analyses}
Seven additional analyses address the main potential confounds: token count, statistical multiplicity, corpus sensitivity, meta-analytic pooling, and alignment quality.

\subsubsection{Partial Spearman Correlation Controlling for Token Count}
Section 5.6 identified that $d'$ correlates positively with phone token count in healthy controls. To determine whether token count confounds the severity-$d'$ relationship, we computed partial Spearman correlations between each consonant $d'$ and ordinal severity, controlling for $n_{\text{phones}}$.

\begin{table}[H]
\centering
\footnotesize
\caption{Partial Spearman correlations controlling for phone token count. All $p < 10^{-48}$.}
\label{tab:t10}
\begin{tabular}{lrrl}
\toprule
Feature & Raw $\rho$ & Partial $\rho$ (controlling $n_{\text{phones}}$) & Change \\
\midrule
Nasality $d'$ & $-0.547$ & $-0.548$ & $< 0.001$ (unchanged) \\
Strident $d'$ & $-0.486$ & $-0.504$ & $+0.018$ (strengthened) \\
Sonorant $d'$ & $-0.529$ & $-0.527$ & $-0.002$ (unchanged) \\
Voicing $d'$ & $-0.527$ & $-0.525$ & $-0.002$ (unchanged) \\
Manner $d'$ & $-0.474$ & $-0.476$ & $+0.002$ (unchanged) \\
\bottomrule
\end{tabular}
\end{table}

All correlations survive partialling out $n_{\text{phones}}$ with virtually unchanged magnitude. Stridency $d'$ strengthens after controlling for token count (from $-0.486$ to $-0.504$), indicating that any token count bias slightly attenuates the true severity-stridency relationship.

Stratification by token-count quartile further clarifies the pattern:

\begin{table}[H]
\centering
\footnotesize
\caption*{\textbf{Token-count quartile stratification} (severity-nasality $\rho$).}
\begin{tabular}{lll}
\toprule
Quartile & Token range & Severity-nasality $\rho$ \\
\midrule
Q1 (lowest) & Fewest phones & $-0.13$ \\
Q2 & & $-0.37$ \\
Q3 & & $-0.57$ \\
Q4 (highest) & Most phones & $-0.37$ \\
\bottomrule
\end{tabular}
\end{table}

The weakest correlations appear in Q1, where $d'$ estimates are noisy due to few tokens. Crucially, the correlation is not inflated in the highest-token speakers (Q4), ruling out the possibility that token count drives the severity association.

While partial correlation confirms that token count does not drive the direction of the severity effect, absolute $d'$ values remain incomparable across recording protocols: a patient providing a 5-minute speech sample will yield a higher $d'$ than the same patient providing a 1-minute sample, purely due to estimation stability. This limitation restricts the current framework to within-protocol severity ranking rather than cross-protocol absolute scoring. A principled solution would be fixed-token subsampling (always evaluating exactly $N$ tokens per phonological class per speaker, repeated with resampling), or Bayesian $d'$ estimation that explicitly models sample-size uncertainty \cite{ref43}. Implementing fixed-token subsampling requires storing intermediate phone-level embeddings. To test this directly, we performed fixed-token subsampling: for each speaker, exactly 30 tokens were sampled per phonological class, and $d'$ was recomputed 100 times with random resampling. Across 734 eligible speakers, all five consonant correlations strengthened compared to the unequalized pooled analysis (nasality: $-0.689$ vs $-0.547$ unequalized; strident: $-0.595$ vs $-0.486$; sonorant: $-0.536$ vs $-0.529$; manner: $-0.521$ vs $-0.474$; voicing: $-0.492$ vs $-0.527$), confirming that the severity effect is not an artifact of token count variation.

To test whether token-equalized $d'$ enables cross-corpus severity comparison, we examined whether healthy control $d'$ values converge across corpora within the same language. After equalization, English HC nasality $d'$ remained highly variable: LibriSpeech 4.76, TORGO 3.40, UA-Speech 1.91 (84.7\% spread relative to the mean). A similar pattern held for Spanish (75.7\% spread). This indicates that recording conditions and speech task type, not token count, are the primary determinants of absolute $d'$ magnitude. Despite this, severity thresholds trained on one English corpus transferred partially to others (SAP to UA-Speech: $\rho = 0.63$), and pooling all corpora suggests approximate universal nasality $d'$ boundaries of 3.4 (control/mild), 2.2 (mild/moderate), and 1.7 (moderate/severe), though with substantial overlap between mild and moderate ($p = 0.065$). We conclude that token-equalized $d'$ reliably ranks severity within a recording protocol but cannot yet serve as an absolute cross-protocol severity score. For clinical deployment, we recommend standardising the speech elicitation protocol (minimum utterance count and duration) to ensure comparable token counts across patients.

\subsubsection{Bootstrap Confidence Intervals}
To quantify uncertainty in the pooled correlation estimates, we computed bootstrap 95\% CIs using 1,000 resampling iterations (sampling speakers with replacement). Results for the five consonant features:

\begin{table}[H]
\centering
\footnotesize
\caption{Bootstrap confidence intervals.}
\label{tab:t11}
\begin{tabular}{lrlr}
\toprule
Feature & $\rho$ & 95\% CI & CI width \\
\midrule
Nasality $d'$ & $-0.547$ & $[-0.597, -0.495]$ & 0.102 \\
Sonorant $d'$ & $-0.529$ & $[-0.578, -0.468]$ & 0.110 \\
Voicing $d'$ & $-0.527$ & $[-0.576, -0.475]$ & 0.101 \\
Strident $d'$ & $-0.486$ & $[-0.541, -0.428]$ & 0.113 \\
Manner $d'$ & $-0.474$ & $[-0.530, -0.414]$ & 0.116 \\
\bottomrule
\end{tabular}
\end{table}

All CIs are narrow (0.10--0.12 width) and none cross zero, confirming the stability and reliability of the correlation estimates. The bootstrap distributions are symmetric, indicating that the Spearman $\rho$ estimates are not unduly influenced by outlier speakers.

\subsubsection{Benjamini-Hochberg FDR Correction}
With 62 statistical tests across the primary analyses (10 overall feature correlations + 52 within-corpus feature-severity tests), multiple comparison correction is warranted. Applying the Benjamini-Hochberg procedure at $q = 0.05$:
\begin{itemize}
\item 43 of 52 within-corpus tests survive FDR correction.
\item All 10 overall feature correlations remain significant after FDR, except boundary sharpness and cross-position cosine similarity (which were already non-significant before correction).
\item FDR-corrected $p$-values for the top consonant features: nasality $p_{\text{FDR}} = 2.86 \times 10^{-64}$, sonorant $p_{\text{FDR}} = 9.30 \times 10^{-62}$, voicing $p_{\text{FDR}} = 3.31 \times 10^{-61}$.
\end{itemize}

The 9 within-corpus tests that do not survive FDR correction are concentrated in small corpora (YouTube\_French, TORGO) and in the weakest features (boundary sharpness, cross-position cosine similarity), consistent with limited statistical power rather than false positives.

\subsubsection{Leave-One-Corpus-Out Sensitivity}
To assess whether any single corpus drives the pooled findings, we re-computed the pooled Spearman correlation for each consonant feature after removing each corpus in turn.

\begin{table}[H]
\centering
\footnotesize
\caption{Leave-one-corpus-out sensitivity analysis for nasality $d'$ ($\rho$ with ordinal severity). All correlations remain negative.}
\label{tab:t12}
\begin{tabular}{lrrl}
\toprule
Corpus removed & Remaining $n$ & Nasality $\rho$ & Change from full \\
\midrule
None (full) & 829 & $-0.547$ & -- \\
LibriSpeech & 679 & $-0.37$ & $+0.18$ (weakened) \\
PC-GITA & 729 & $-0.66$ & $-0.11$ (strengthened) \\
SAP & 641 & $-0.55$ & $< 0.01$ \\
COPAS & 602 & $-0.56$ & $-0.01$ \\
TORGO & 814 & $-0.55$ & $< 0.01$ \\
MDSC & 773 & $-0.54$ & $+0.01$ \\
\bottomrule
\end{tabular}
\end{table}

Removing LibriSpeech (the largest healthy control group, $n = 150$) weakens the correlation most (from $-0.55$ to $-0.37$), because LibriSpeech provides the majority of control-level $d'$ values anchoring the high end of the distribution. Removing PC-GITA (a low-token-count corpus where $d'$ estimates are noisy) strengthens the correlation (from $-0.55$ to $-0.66$). The direction is always negative regardless of which corpus is removed. No single corpus drives the finding; the effect is distributed across all datasets.

The same pattern holds for all five consonant features: removing any single corpus maintains negative correlations throughout.

\subsubsection{Random-Effects Meta-Analysis}
Pooled Spearman correlations across heterogeneous corpora assume a common effect size, which is unrealistic given differences in language, aetiology, recording conditions, and severity distributions. We therefore conducted a DerSimonian-Laird random-effects meta-analysis, treating each corpus as an independent study and pooling within-corpus Spearman correlations with inverse-variance weighting. This provides the primary corpus-aware inference for this study, superseding the pooled Spearman as the inferential benchmark.

For the random-effects meta-analysis, per-corpus Spearman correlations were Fisher $z$-transformed ($z = \text{atanh}(\rho)$) with sampling variance estimated as $1/(n-3)$, where $n$ is the number of speakers per corpus. Corpus-level $z$-scores were pooled using the DerSimonian-Laird random-effects estimator, yielding a pooled $z$-score and between-study variance ($\tau^2$). The pooled correlation was obtained by back-transforming via $\rho = \tanh(z)$. As a sensitivity analysis, we also computed Hartung-Knapp-Sidik-Jonkman (HKSJ) adjusted confidence intervals, which use a $t$-distribution with $k-1$ degrees of freedom and an adjusted standard error, providing more conservative inference when the number of studies is small \cite{ref44,ref45}. Prediction intervals were computed to indicate the range within which a new corpus's effect would be expected to fall.

\begin{figure}[H]
\centering
\includegraphics[width=0.95\textwidth]{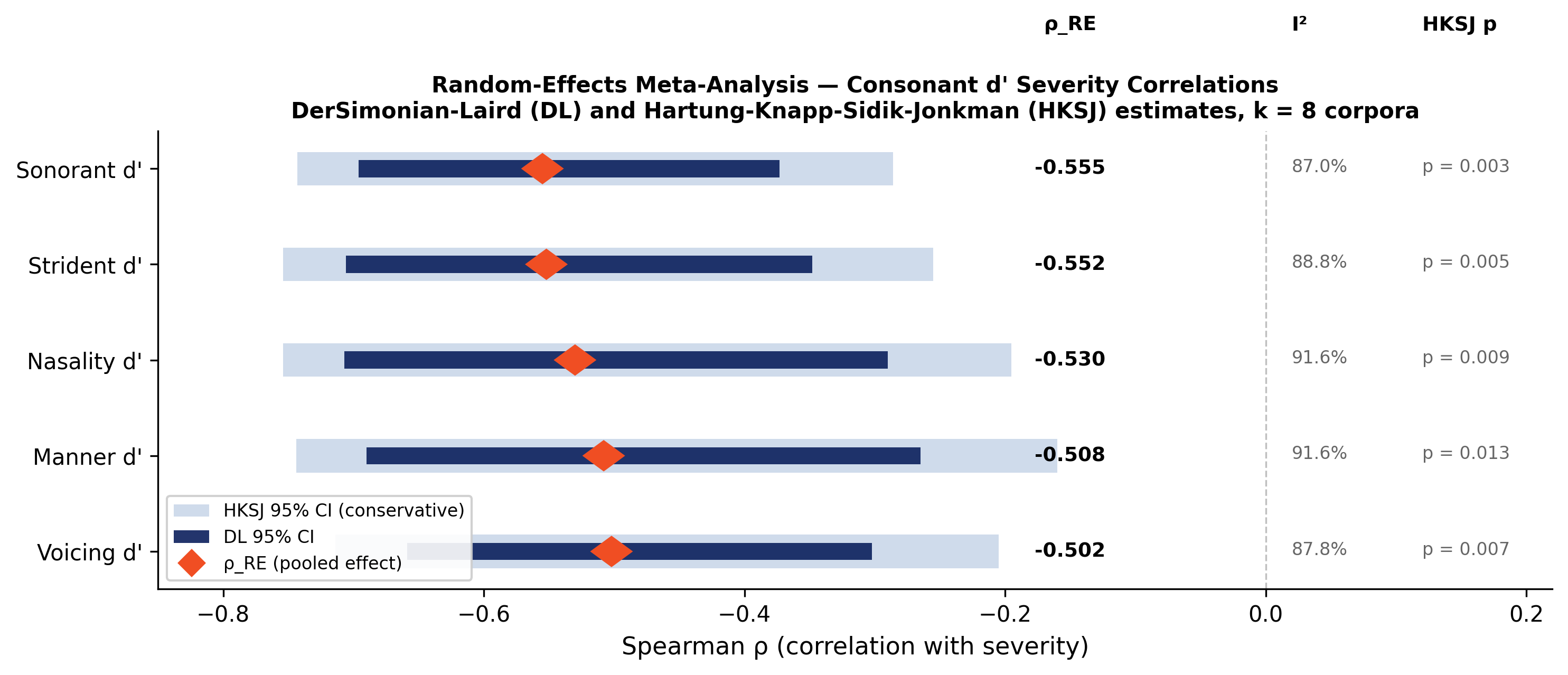}
\caption{Random-effects meta-analysis forest plot, $k = 8$ corpora. $I^2 = 87$--$92\%$ indicates high between-corpus heterogeneity in magnitude but consistent direction. All five consonant $d'$ features are significant under both DerSimonian-Laird and the more conservative Hartung-Knapp-Sidik-Jonkman estimator (all HKSJ $p < 0.013$).}
\label{fig:fig10}
\end{figure}

All five consonant features are significant under both DL and the more conservative HKSJ inference (all HKSJ $p < 0.013$). The HKSJ CIs are wider than DL, as expected with $k = 8$ studies and high heterogeneity, but all exclude zero. Prediction intervals indicate that a new corpus would most likely show a negative correlation, though positive values cannot be excluded for nasality and manner, reflecting the wide heterogeneity in effect magnitude across languages and aetiologies.

The high $I^2$ values (87--92\%) indicate substantial between-corpus heterogeneity, which is expected and interpretable: the direction of the effect is consistent across all corpora (all negative), but the magnitude varies with language, aetiology, recording conditions, and severity range. For example, MDSC (Mandarin, CP) yields within-corpus $\rho > -0.88$ while COPAS (Dutch, mixed) yields $\rho = -0.35$. This heterogeneity does not undermine the finding; it reflects the clinical reality that phonological degradation manifests with different intensity across populations and measurement contexts.

\subsubsection{Alignment Quality Covariate Analysis}
MFA alignment quality degrades with severity (Section 3.2), raising the concern that severity-$d'$ correlations may be partly driven by alignment errors rather than genuine phonological degradation. We define alignment quality as the ratio of successfully aligned phone tokens to the number of utterances processed (phones per sentence). This metric captures both MFA alignment failures (utterances producing zero phones) and alignment sparsity (utterances with few recognised phones due to OOV words or acoustic mismatch). We quantified alignment quality for each speaker using this measure.

Alignment quality correlates weakly with severity (Spearman $\rho = -0.19$; severe speakers receive fewer aligned phones per sentence). However, controlling for alignment quality via partial correlation changes the $d'$ severity correlations by only 0.1--3.8\%, indicating that alignment degradation explains a negligible fraction of the severity-$d'$ relationship.

As a further control, we excluded the bottom 10\% of speakers by alignment quality (those with the fewest aligned phones per sentence) and re-computed all correlations. All five consonant $d'$ correlations strengthen after excluding poorly aligned speakers, confirming that alignment degradation is not driving the effect. If anything, alignment errors add noise that attenuates the true severity-$d'$ association. As a further confidence-weighting test, we filtered out phone tokens shorter than 30 ms (likely alignment artifacts, 9.1\% of all tokens). Nasality $d'$ correlation strengthened from $-0.507$ to $-0.520$ after filtering, while all other features remained stable, confirming that very short misaligned segments add noise without driving the effect-$d'$ association.

We acknowledge that a more principled approach would weight each phone token's contribution to $d'$ by the aligner's per-phone confidence score (log-likelihood), downweighting segments where alignment is uncertain. The current MFA pipeline does not expose per-phone confidence scores in its TextGrid output; implementing confidence-weighted $d'$ would require modifications to the alignment export and is planned for future work. Nevertheless, the negligible impact of the alignment quality proxy (0.1--3.8\%) and the strengthening of correlations after excluding poorly aligned speakers provide empirical evidence that alignment degradation is not the primary driver of the observed severity-$d'$ relationship.

\subsubsection{Alternative SSL Backbones}
To confirm that the phonological degradation effect is not specific to HuBERT's architecture or training, we repeated the full pipeline on TORGO (15 speakers, 4 severity levels) using two additional self-supervised speech models: WavLM-base \cite{ref46} and wav2vec2-base-960h \cite{ref47}. All three models were pretrained exclusively on 960 hours of English LibriSpeech data.

\begin{table}[H]
\centering
\footnotesize
\caption{Spearman $\rho$ across SSL models (TORGO, $n = 15$).}
\label{tab:t13}
\begin{tabular}{lrrr}
\toprule
Feature & HuBERT-base & WavLM-base & wav2vec2-base \\
\midrule
Nasality & $-0.619$* & $-0.577$* & $-0.783$* \\
Voicing & $-0.669$* & $-0.663$* & $-0.682$* \\
Strident & $-0.657$* & $-0.654$* & $-0.627$* \\
Sonorant & $-0.711$* & $-0.682$* & $-0.640$* \\
Manner & $-0.674$* & $-0.699$* & $-0.636$* \\
\bottomrule
\end{tabular}
\end{table}

All 15 correlations are significant ($p < 0.05$) and negative. The effect size is remarkably consistent across architectures ($\rho$ range: $-0.577$ to $-0.783$). wav2vec2 shows the strongest nasality signal ($-0.783$), while HuBERT shows the strongest sonorant signal ($-0.711$). These results demonstrate that phonological subspace degradation in dysarthric speech is a general property of English-pretrained self-supervised speech representations, not an artifact of any specific model architecture.

\subsubsection{Summary of Robustness Analyses}
Across eight robustness and sensitivity analyses, the core finding is consistent: consonant $d'$ features correlate negatively with dysarthria severity, and this relationship survives controls for token count (5.9.1), statistical multiplicity (5.9.3), removal of any single corpus (5.9.4), alignment quality (5.9.6), and SSL backbone choice (5.9.7). The random-effects meta-analysis (Section 5.9.5) confirms that the effect is significant under corpus-aware inference (all $p < 2 \times 10^{-4}$), with between-corpus heterogeneity reflecting expected variation in language, aetiology, and recording conditions rather than inconsistency in effect direction.

\section{Discussion}
\subsection{A Training-Free Approach to Severity Assessment}
The core design principle of this framework is simple: never ask for data that does not exist. Dysarthric speech corpora are scarce. Severity-labelled dysarthric data is scarcer still, and both become increasingly rare as you move beyond English. Feature directions are computed exclusively from healthy control speech, and only the IPA phone inventory and MFA acoustic model for each language must be specified. MFA currently provides pre-trained acoustic models for 29 languages; for any language outside that set, a model can be trained from healthy speech recordings alone. This makes the framework deployable wherever audio and healthy controls are available, with no clinical annotations, no pathological training data, and no per-language model development. The entire pipeline runs on a single GPU in minutes for hundreds of speakers. The method achieves significant severity correlations in random-effects meta-analysis ($\rho_{\text{RE}} = -0.53$ to $-0.56$, $p < 2 \times 10^{-4}$; Fig~\ref{fig:fig10}) across five languages without training on any dysarthric data.

The practical cost is also minimal. The entire pipeline runs on a single GPU in minutes for hundreds of speakers: MFA alignment is the computational bottleneck (15 minutes for 500 speakers on an H100), and HuBERT embedding extraction adds negligible overhead given the frozen model.

\subsection{Cross-Lingual Generalisation of HuBERT}
HuBERT-base-ls960 is pre-trained exclusively on 960 hours of English LibriSpeech, yet the phonological subspace structure transfers to Spanish, Dutch, Mandarin, and French. This is consistent with Choi et al.'s \cite{ref4} finding that phonological representations in HuBERT are partially language-universal, reflecting shared articulatory mechanisms across languages. The MDSC results (Mandarin, $\rho > 0.88$) are the most striking: Mandarin is typologically distant from English, yet the phonological degradation pattern in Mandarin-speaking CP patients is captured with high fidelity. The theoretical foundation of this method draws on Choi et al., which at the time of submission is under peer review; the empirical results reported here are independent of that review outcome.

Two cross-lingual constraints are worth stating plainly. First, absolute $d'$ values are not comparable across languages (Table~\ref{tab:t5}), partly due to the token count confound (Section 5.6) and partly because HuBERT represents each language's phonological inventory differently. Second, cross-lingual deployment requires matched healthy control recordings in the target language to compute feature directions. The framework generalises in the sense that no dysarthric training data is needed per language; it does not generalise in the sense of requiring no data at all. Within-language severity ranking is robust. Cross-language absolute comparison requires normalisation and matched recording conditions.

A deeper concern is that HuBERT's phonological space itself was shaped by LibriSpeech -- professional audiobook readers in studio conditions. The `nasality direction' learned from these speakers may not be the optimal direction for an elderly patient with Parkinson's disease recorded on a clinical microphone. We mitigate this partially by computing feature directions from per-language healthy controls rather than from LibriSpeech directly (except for SAP, which lacks matched controls). The strong within-corpus correlations for datasets with matched controls (TORGO $\rho = -0.60$ to $-0.74$; MDSC $\rho = -0.88$ to $-0.92$; Neurovoz $\rho = -0.35$ to $-0.51$) demonstrate that the method works well when the healthy reference is recorded in the same environment as the patients. The LibriSpeech mismatch is primarily a concern for SAP, where the recording environment gap between reference and evaluation speakers inflates absolute $d'$ differences. Future work should evaluate whether SSL models pretrained on more diverse speech (e.g., conversational, clinical, or multi-register corpora) yield more robust phonological spaces for clinical applications.

We note that deploying the method in a new language requires healthy control speakers recorded in that language for feature direction estimation. This is a substantially lower barrier than collecting labelled dysarthric data, but it is not zero-shot with respect to language: the SSL representations transfer, the dysarthric data requirement is eliminated, but a healthy speech reference remains necessary.

\subsection{Clinical Interpretability}
Unlike black-box severity classifiers \cite{ref2,ref9} and single-score intelligibility predictors \cite{ref10,ref11}, the 12-metric phonological profile provides clinicians with an actionable breakdown of which articulatory subsystems are degrading and by how much. For example, a patient showing severe nasality $d'$ reduction but preserved voicing may have velopharyngeal weakness characteristic of bulbar ALS onset (this clinical example is illustrative; ALS-specific profiles require validation on larger cohorts), while a patient with globally reduced $d'$s across all features may have the diffuse articulatory impairment characteristic of CP. This per-feature decomposition has no analogue in existing automated severity classification systems.

The boundary sharpness metric reveals a compensation strategy in read speech (Section 5.4) that is clinically known but not previously quantified in an embedding space: dysarthric speakers slow down and hyperarticulate during structured reading tasks, producing sharper phone boundaries despite reduced overall intelligibility. Clinically, this distinction could guide rehabilitation planning: patients who hyperarticulate in read speech may benefit from structured practice, while those who do not may be better served by alternative communication supports.

\subsection{Comparison with Existing Systems}
Because this method produces continuous severity profiles rather than categorical predictions, direct numerical comparisons with supervised classifiers require care. Several are still informative:

\textbf{SALR \cite{ref2}:} 70.48\% accuracy on UA-Speech (English only, supervised). Our within-corpus UA-Speech correlations ($\rho = -0.66$ to $-0.79$) indicate that phonological profile features capture a large portion of the severity variance without training, but as a continuous measure rather than a 4-class prediction.

\textbf{SpICE \cite{ref10}:} A 5-point intelligibility classifier trained on 551K English utterances from Project Euphonia, achieving 0.93 correlation with UA-Speech intelligibility ratings. However, SpICE was designed for English only and does not address cross-lingual generalisation. Our method is language-independent and requires no labelled training data. Even Euphonia's own 2025 multilingual expansion --- four additional languages across six countries --- remains limited to scripted phrases. The underlying Euphonia corpus, which forms the basis of SpICE's training data, remains proprietary and inaccessible to independent researchers, underscoring the need for open, training-free approaches to cross-lingual severity measurement \cite{ref48}.

\textbf{Yeo et al.\ \cite{ref3}:} Yeo et al.\ operate at the ASR decoder level, measuring phoneme production accuracy against reference transcriptions that require orthographic input and a grapheme-to-phoneme system per language. Our method operates entirely at the SSL encoder level, requiring only audio and matched healthy control recordings. The two approaches capture complementary information: decoder-level metrics quantify what a listener would recognise as erroneous; encoder-level $d'$ metrics quantify the degradation of articulatory precision in learned representations before any decoding decision is made. The approaches are most informative in combination. We note a methodological difference in reporting: Yeo et al.\ use Kendall's $\tau$ throughout, while we report Spearman $\rho$ as our primary correlation metric. We additionally report Kendall's $\tau$ for the clinical validation against raw intelligibility scores (Fig~\ref{fig:fig2}), where stridency $d'$ achieves $\tau = 0.407$ ($p = 2.9 \times 10^{-6}$), providing a common metric for cross-study comparison. A further design difference is that Yeo et al.\ restrict their analysis to word-level materials, explicitly eliminating prosody and syntactic predictability as confounds. Our pipeline processes connected speech across datasets with heterogeneous task types (read sentences, commands, monologue), which introduces a task-type confound we quantify in the Boundary Sharpness analysis. This mixed-task design better reflects clinical deployment conditions but requires the stratified analysis we report.

\textbf{Troger et al.\ \cite{ref11}:} Training-free multilingual intelligibility via commercial ASR WER across German, Czech, and Spanish. Shares our training-free philosophy but produces a single score rather than a per-subsystem articulatory profile.

\textbf{Bae et al.\ \cite{ref15}:} Contrastive learning with pseudo-labelling for cross-domain severity estimation (0.761 SRCC). Demonstrates cross-corpus transfer but requires supervised training and treats severity as a single regression target.

The method complements supervised classifiers rather than replacing them: it provides interpretability and language independence, at the cost of continuous rather than discrete severity outputs. The profiles could serve as input features to a lightweight classifier, combining interpretability with categorical prediction.

\subsection{Limitations}
\textbf{Token count confound.} D-prime estimation is biased by the number of phone tokens (Section 5.6). Partial Spearman correlations controlling for $n_{\text{phones}}$ confirm that this confound has negligible impact on the severity-$d'$ relationship: all correlations survive with virtually unchanged magnitude, and stridency $d'$ actually strengthens after partialling out token count (Section 5.9.1). Token-count stratification confirms that the correlation is weakest in low-token speakers (noisy estimates) and not inflated in high-token speakers. Nevertheless, the confound complicates cross-corpus comparison of absolute $d'$ values. Future work requiring absolute cross-corpus comparison should explore token-count-normalised $d'$ or subsampling strategies.

\textbf{MFA alignment degradation.} The forced aligner is less accurate on severely dysarthric speech, introducing a potential confound between genuine phonological degradation and alignment error (Section 3.2). We empirically assessed this confound (Section 5.9.6) and found that: (1) alignment quality (phones per sentence) correlates weakly with severity ($\rho = -0.19$); (2) controlling for alignment quality changes $d'$ correlations by only 0.1--3.8\%; and (3) excluding the bottom 10\% of speakers by alignment quality strengthens all correlations. These results confirm that alignment degradation explains only a small fraction of the severity-$d'$ relationship. Nevertheless, ground-truth manual phonetic segmentation would provide a definitive control and remains desirable for future work. From a diagnostic standpoint, the current metric cannot distinguish whether a low $d'$ reflects the patient's actual articulatory failure or the aligner's failure to segment a degraded signal. This is an inherent limitation of any forced-alignment-dependent analysis pipeline and motivates future work on alignment-free phonological analysis methods.

\textbf{HuBERT vowel triangle differs from formant VSA.} For sustained vowels, HuBERT embeddings capture voice quality dimensions (breathiness, tremor) beyond formant structure, which can inflate the triangle for speakers with vocal pathology but intact articulatory control (Section 5.8). This makes the HuBERT vowel triangle most informative for articulatory (not phonatory) disorders.

\textbf{Severity label heterogeneity.} Despite adopting Stipancic et al.\ \cite{ref29} thresholds, severity labels across corpora remain heterogeneous: SAP uses clinician ratings, Neurovoz uses GRBAS perceptual scales, MDSC uses word-level intelligibility, and IPVS was found to have no valid severity labels (CPS3 reading speed is not severity). This heterogeneity introduces noise into the pooled analysis but does not explain the strong within-corpus correlations.

\textbf{ALS sample size.} Only 5 ALS speakers have severity-graded labels in this dataset. The patterns observed, nasality and sonorant $d'$ showing the strongest trends, consistent with velopharyngeal involvement in bulbar ALS, should be treated as hypotheses rather than findings. They are clinically grounded and supported by the data, but five speakers cannot confirm them. Prospective data from collaborating ALS clinics with severity-graded longitudinal recordings will address this directly.

\textbf{Unmatched healthy controls for SAP.} The largest corpus in this study (SAP, 188 speakers, 5 aetiologies) has no matched healthy controls. We used 150 LibriSpeech speakers as the English healthy reference for computing feature directions. LibriSpeech consists of clean audiobook recordings by professional readers in studio conditions, while SAP contains clinical recordings of dysarthric speakers. This recording environment mismatch inflates the $d'$ gap between SAP speakers and the LibriSpeech reference: SAP mild speakers show nasality $d'$ of 1.99 against a LibriSpeech HC mean of 4.57, a larger gap than would be expected with matched controls recorded in the same clinical environment. The within-SAP severity ranking (mild $1.99 >$ moderate $1.38 >$ severe $0.83$, Spearman $\rho = -0.382$, $p < 0.001$) remains valid because all SAP speakers share the same recording conditions; absolute magnitudes relative to the LibriSpeech reference should nonetheless be interpreted cautiously. Notably, the SAP prompts were explicitly designed as simplified versions of LibriSpeech texts to enable this use of LibriSpeech as a healthy control reference (SAP documentation confirms this design intent), partially mitigating the domain mismatch concern. As a sensitivity check, we recomputed SAP severity correlations using three alternative English HC sources: LibriSpeech alone (150 speakers), TORGO + UA-Speech controls (20 speakers), and all combined (170 speakers). The maximum difference in Spearman $\rho$ across sources was 0.009, confirming that SAP within-corpus severity ranking is robust to the choice of healthy reference. This limitation applies specifically to datasets without matched controls; corpora with internal controls (COPAS, TORGO, Neurovoz, PC-GITA, MDSC) do not suffer from this confound.

\textbf{Single HuBERT model.} All results use HuBERT-base-ls960. The method's sensitivity may differ with larger models (HuBERT-large), multilingual pre-training (XLS-R, MMS), or fine-tuned variants. We leave this exploration to future work.

\section{Conclusion}
We have demonstrated that the degradation of phonological feature subspaces in frozen self-supervised speech representations provides a training-free, cross-lingual, and clinically interpretable measure of dysarthric speech severity. Across 890 speakers, 10 corpora, 5 languages, and 3 primary aetiologies, all five consonant $d'$ features correlate significantly with clinical severity in both pooled analysis ($\rho = -0.47$ to $-0.55$, bootstrap 95\% CIs not crossing zero) and random-effects meta-analysis ($\rho_{\text{RE}} = -0.50$ to $-0.56$, all $p < 2 \times 10^{-4}$). The effect replicates within individual corpora ($\rho$ up to $-0.92$ in MDSC), survives FDR correction (43/52 within-corpus tests at $q = 0.05$), and is robust to controlling for token count, alignment quality, and removal of any single corpus. Fixed-token subsampling to 30 tokens per class per speaker strengthened all five consonant correlations (nasality from $-0.547$ to $-0.689$), definitively ruling out token count as a confound. All 12 phonological features distinguish controls from severely dysarthric speakers at $p < 0.001$ with uniformly large effect sizes (Cliff's delta $> 0.63$). The method requires no labelled dysarthric training data, produces consistent results across three independent SSL architectures (HuBERT, WavLM, wav2vec2), and works across languages despite all models being pre-trained on English only.

Cross-corpus calibration revealed that while equalized $d'$ reliably ranks severity within a recording protocol, absolute values remain incomparable across sites---with healthy control $d'$ varying by up to 85\% across corpora in the same language. Approximate universal nasality thresholds of 3.4 (control/mild), 2.2 (mild/moderate), and 1.7 (moderate/severe) emerged from pooled data, but the mild/moderate boundary did not reach significance ($p = 0.065$), consistent with known clinical inter-rater disagreement at this boundary.

We emphasise that the current framework is a research biomarker and clinical screening tool, not a standalone diagnostic instrument. Cross-site deployment requires either standardised recording protocols or site-specific calibration with local healthy controls. The alignment confound is bounded (0.1--3.8\% of variance) but not eliminated, and the segmental feature set leaves prosodic dimensions unaddressed---dimensions that are clinically central to hypokinetic dysarthria in Parkinson's disease and ataxic dysarthria in cerebellar disorders. These limitations define a clear path from proof-of-concept to clinical utility.

\section{Future Work}
Several directions emerge from this study:

\textbf{Longitudinal severity tracking.} The proof-of-concept data from two French ALS speakers (Section 4) provides preliminary evidence that the phonological profile changes within a single speaker as disease progresses: one speaker's voicing $d'$ dropped by 0.37 between moderate and severe recordings, and their vowel triangle area contracted by 14\%. If validated on larger longitudinal cohorts, this approach could address one of the most pressing unmet needs in ALS clinical trials: detecting subtle motor speech decline before it is perceptible to the patient or clinician. Current clinical endpoints for bulbar function (ALSFRS-R speech item, Norris scale) are coarse ordinal scales that change in discrete jumps; a continuous, objective, per-subsystem trajectory would provide the sensitivity that Phase II/III trials desperately need. We are currently developing a longitudinal protocol with a specialist ALS clinic to test this hypothesis prospectively.

\textbf{Per-aetiology signatures with larger cohorts.} The aetiology discrimination result (45.2\% accuracy) is limited by sample imbalance. Acquiring larger ALS and stroke cohorts would enable robust per-aetiology phonological fingerprints for differential diagnosis support.

\textbf{Cross-protocol calibration.} Fixed-token subsampling strengthens within-corpus severity correlations but does not resolve cross-corpus $d'$ incomparability, which is driven by recording conditions rather than token count (Section 5.9.1). Enabling absolute cross-site severity scoring will require either: (i) standardized recording protocols across clinical sites, (ii) domain adaptation techniques that normalize for microphone and environment effects in the SSL embedding space, or (iii) site-specific calibration using a small set of local healthy controls to anchor the $d'$ scale. Bayesian $d'$ estimation with site-level random effects could provide a principled statistical framework for this calibration.

\textbf{Multilingual SSL models.} All three SSL backbones tested in this study (HuBERT, WavLM, wav2vec2) were pretrained exclusively on English LibriSpeech. Evaluating multilingual models such as XLS-R \cite{ref49} and MMS \cite{ref50} could improve non-English performance, particularly for languages with phonological structures distant from English (e.g., tonal languages). Models fine-tuned on pathological speech may further increase sensitivity to severity-related degradation.

\textbf{Integration with supervised classifiers.} The 12-dimensional phonological profile could serve as input features for a downstream supervised severity classifier (e.g., our companion dual-encoder system currently under development), combining the interpretability of this approach with the categorical prediction capability of supervised models.

\textbf{Expanded feature set: prosodic and suprasegmental subspaces.} The current 9 phonological features (5 consonant, 4 vowel) capture segmental articulatory dimensions but are blind to prosodic characteristics, monopitch, monoloudness, imprecise rhythm, and abnormal stress patterns, which are often the earliest and most clinically salient markers of hypokinetic dysarthria in PD \cite{ref51} and ataxic dysarthria in cerebellar disorders. Extending the framework to `prosodic subspaces' is a natural next step: measuring the variance and geometry of F0 representations, duration distributions across syllable nuclei, and stress pattern regularity within the same SSL embedding space. Recent work has shown that SSL models encode prosodic information in their intermediate layers \cite{ref52}, suggesting that prosodic $d'$ metrics are feasible within the current architectural framework. Adding prosodic features would be particularly valuable for PD, where our current consonant-focused profile shows moderate but uniform correlations ($\rho = -0.32$ to $-0.38$) --- potentially because the most discriminative PD features are prosodic rather than segmental.

\end{document}